\documentclass[10pt,twocolumn,letterpaper]{article}

\usepackage[pagenumbers]{wacv} % To force page numbers, e.g. for an arXiv version

\usepackage{pifont}
\usepackage{multirow}
\usepackage{wrapfig}
\usepackage{makecell}
\usepackage{booktabs}
\usepackage{graphicx}
\usepackage{subcaption}
\usepackage[table]{xcolor}

\usepackage[utf8]{inputenc}
\usepackage[T1]{fontenc}

\usepackage{url}
\usepackage{amsfonts}
\usepackage{amsmath}
\usepackage{nicefrac}
\usepackage{microtype}

\usepackage{algorithm}
\usepackage{algorithmicx}
\usepackage{algpseudocode}

\newcommand{\ourcell}[1]{\cellcolor{blue!8}#1}

\definecolor{wacvblue}{rgb}{0.21,0.49,0.74}
\usepackage[pagebackref,breaklinks,colorlinks,allcolors=wacvblue]{hyperref}

\def\wacvPaperID{1325} % *** Enter the WACV Paper ID here
\def\confName{WACV}
\def\confYear{2027}

\title{Efficient Memory-Constrained Video Object Segmentation via Subspace Optimization}

\author{
Hai-Anh Le$^{*1}$ \qquad
Le-Trung Nguyen$^{*1}$ \qquad
Enzo Tartaglione$^{1,2}$ \\
$^{1}$LTCI, Télécom Paris, Institut Polytechnique de Paris, France \quad $^{2}$CNR-ISASI, Italy\\
\small $^*$Equal contribution\\\texttt{enzo.tartaglione@telecom-paris.fr}
}

\begin{document}
\maketitle

\begingroup
\renewcommand\thefootnote{}
\footnotemark\footnotetext{This article has been accepted for publication at the 2027 IEEE/CVF Winter Conference on Applications of Computer Vision (WACV 2027).}
\endgroup

\begin{abstract}
Memory-based architectures for video object segmentation (VOS) have achieved remarkable performance by maintaining dense feature representations from past frames in a streaming Memory Bank. However, storing these features over time incurs substantial memory overhead and computational cost. In this work, we identify significant temporal redundancy in memory representations, where consecutive frames exhibit highly overlapping low-rank subspaces
that remain stable over time. Building on this observation, we propose Memory Subspace Iteration (MSI), a training-free, model-agnostic plugin that exploits temporal subspace stability to efficiently compress streaming memory features. Extensive experiments on standard VOS benchmarks demonstrate that MSI preserves segmentation accuracy across multiple architectures while reducing the memory bank footprint by up to 25\%. Our implementation is available at \url{https://github.com/Hai-Anh-LE/MSI_WACV2027}.
\end{abstract}
    
\section{Introduction}
\label{sec:intro}
Video Object Segmentation (VOS)~\cite{pont20172017} plays a crucial role in a wide range of domains, including autonomous driving~\cite{camarena2025ad}, healthcare~\cite{zhu2024medical, liu2026unisurgsam}, mobile perception~\cite{zhang2023faster,zhao2023fast}, 3D human-centric vision~\cite{gao2025sam}, and robotics~\cite{yu2025sam}. In many real-world scenarios, deploying these systems requires operating on resource-constrained hardware, minimizing latency, and ensuring data privacy. However, modern large-scale VOS architectures~\cite{cheng2022xmem, ravi2024sam2} pose deployment challenges due to their substantial computational demands and memory footprints.

\begin{figure}[t]
  \centering
  \includegraphics[width=0.48\textwidth]{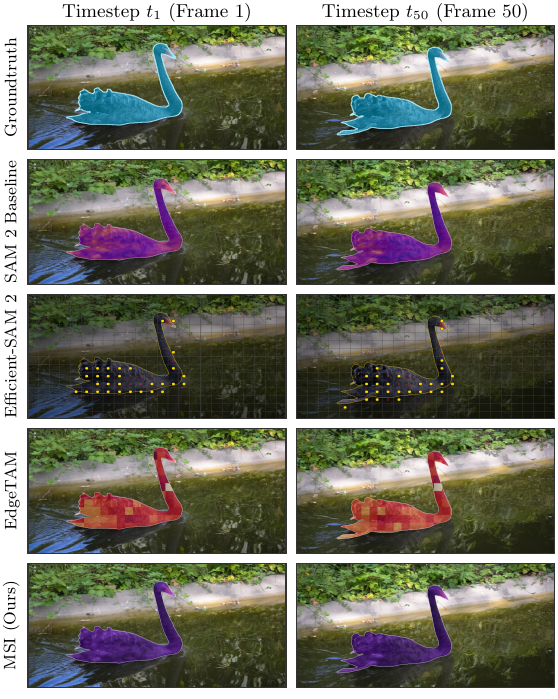}
  \vspace{-2mm}
  \caption{Comparison of memory bank representations at timesteps $t_1$ and $t_{50}$. 
  While token selection (Efficient-SAM~2) discards spatial sampling locations and summary queries (EdgeTAM) blur fine spatial resolution, MSI (Ours) truncates redundant channel dimensions, preserving dense spatial details throughout video propagation.}
  \label{fig:memory_comparison_short}
  \vspace{-3mm}
\end{figure}

Early semi-supervised frameworks, such as XMem~\cite{cheng2022xmem}, DEVA \cite{cheng2023tracking}, and Cutie~\cite{cheng2024putting}, achieved effective long-term tracking using memory-based representations. Subsequently, large vision foundation models significantly improved performance. For example, the Segment Anything Model (SAM)~\cite{kirillov2023segment} was introduced as a strong baseline for static image segmentation. Expanding on this foundation, SAM2~\cite{ravi2024sam2} adapts to video tasks by introducing a streaming Memory Bank and memory attention. In detail, SAM2 uses a memory encoder to combine prior frame representations with object pointers generated by the mask decoder to construct a Memory Bank. Memory attention then retrieves a subset of the Memory Bank for current frames. SAM2 outperformed previous methods in large-scale VOS experiments~\cite{ravi2024sam2}. Recent advances, such as SAM2Long \cite{ding2025sam2longenhancingsam2} and SAM-MT~\cite{shen2026sam-mt}, have extended the capabilities of SAM2 for longer videos and complex scenarios. The latest SAM3~\cite{carion2026sam3} continues to utilize this memory streaming mechanism. We observe that the accumulated memory bank becomes a significant memory bottleneck, as its size grows linearly with the number of processed frames. Thus, it becomes excessively large for long videos. Additionally, retrieving large subsets of these features imposes high computational demands. Together, these limitations hinder deployment in resource-constrained environments.

To address these challenges, existing efficiency-oriented methods generally require additional training and fall into two main categories~\cite{sun2026efficientvariantssegmentmodel}. The first modifies the architecture, with EdgeTAM~\cite{zhou2025edgetam} as a representative approach that trains a 2D Spatial Perceiver to compress memory tensors. The second accelerates memory attention, exemplified by EfficientSAM2~\cite{zhang2026efficient}, which selects object-aware memory tokens for faster retrieval. However, both directions still require training, while EfficientSAM2 additionally retains the dense Memory Bank alongside a sparse one. These limitations motivate a training-free approach that directly reduces memory usage while preserving segmentation performance~\cite{jiaxing2025sam2}.

In this work, we propose \underline{M}emory \underline{S}ubspace \underline{I}teration (MSI), a  training-free framework designed to reduce the memory footprint of streaming segmentation architectures. Our approach relies on the observation that memory features exhibit significant channel-wise redundancy. Specifically, a small subset of dimensions accounts for most of the variation in the memory tensor. Furthermore, we demonstrate that these low-rank subspaces remain stable over time. This allows MSI to efficiently compress the Memory Bank. MSI significantly reduces the storage requirements without architectural changes or fine-tuning. Finally, we design a new retrieval framework that handles these low-rank components and avoids excessive computational overhead.

\noindent{Our main contributions are summarized as follows:}
\begin{itemize}
    \item We show that streaming video segmentation models, such as SAM2 and its variants, suffer from linear growth in memory footprint in Sec.~\ref{sec:problem_statement}
    \item We propose MSI, a training-free framework that leverages the low-rank structure of memory features for efficient compression in Sec.~\ref{sec:msi}. Moreover, we introduce a new low-rank retrieval framework in Sec.~\ref{sec:low-rank_retrieval}.
    \item We conduct extensive experiments in Sec.~\ref{sec:experiments} to illustrate that MSI reduces the Memory Bank size and power consumption while preserving tracking accuracy.
\end{itemize}

\definecolor{lightblue}{RGB}{170,205,235}

\begin{figure*}[t]
    \centering
    \includegraphics[
        width=1.0\linewidth,
        trim=0mm 60mm 0mm 40mm,
        clip
    ]{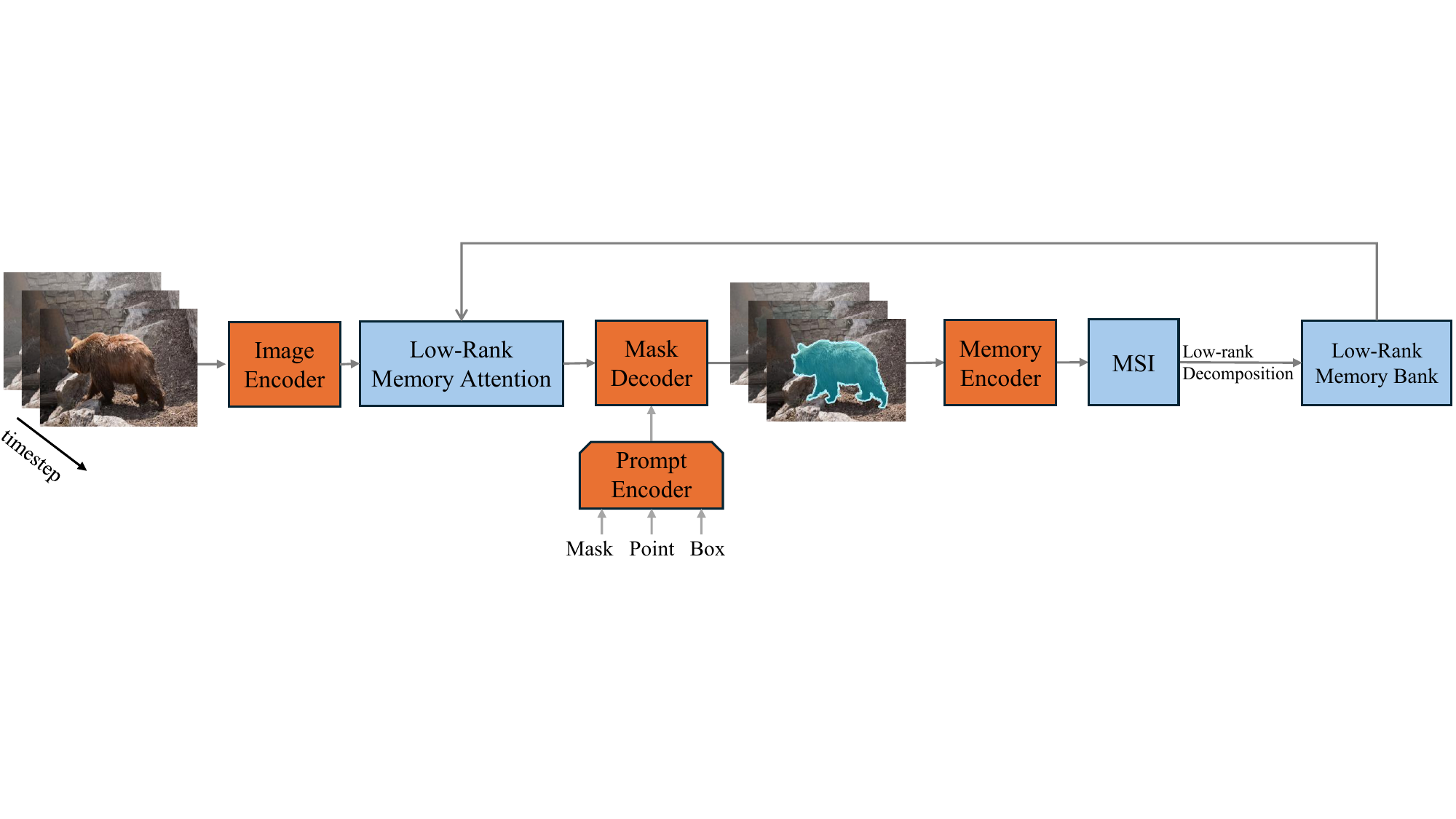}
    \vspace{-2mm}
    \caption{Overview of the MSI framework. MSI compresses incoming memory tensors into a compact low-rank memory bank via subspace iteration, enabling direct low-rank memory attention retrieval without reconstructing dense tensors. \colorbox{lightblue}{Blue} highlights our proposed MSI component, while \colorbox{orange}{orange} indicates the original SAM2 modules.}
    
    \label{fig:msi_pipeline}
    \vspace{-2mm}
\end{figure*}

\section{Related Work}

\paragraph{Segment Anything Model.} Vision foundation models such as SAM~\cite{kirillov2023segment, zhang2023mobilesamv2, zhao2023fast} and its video-centric successor SAM~2~\cite{ravi2024sam2} enable promptable segmentation across images and videos. SAM~2 maintains a streaming Memory Bank and achieves strong performance in Video Object Segmentation~\cite{ding2023mose, hong2023lvos, zhang2026advancingcomplexvideoobject, youtube}, with applications in robotics~\cite{yu2025sam, fang2025sam2act} and medical imaging~\cite{zhu2024medical, yang2025sam2, liu2025sam2s}. More recently, SAM~3~\cite{carion2026sam3}, continues to use this streaming memory mechanism. However, this design relies on a dense Memory Bank and a costly retrieval phase, leading to high storage and latency costs~\cite{xiong2025efficient}, which limits deployment on resource-constrained devices~\cite{lin2023tiny}.

\paragraph{Accelerating Video Models.} Existing approaches to accelerate SAM2 follow two main directions. Architectural redesign methods, such as EdgeTAM~\cite{zhou2025edgetam}, improve efficiency by modifying the model architecture to reduce the cost of inference. EdgeTAM adopts knowledge distillation and trains a 2D Spatial Perceiver to compress memory features into compact global summaries and spatially structured local patches. While promising, these methods require costly distillation and pretraining. Moreover, they need access to downstream data, which limits their robustness and scalability in widely diverse settings.

The second approach focuses on accelerating memory attention. A state-of-the-art method in this direction, EfficientSAM2~\cite{zhang2026efficient}, aims to reduce latency through its Sparse Memory Retrieval module, which selects salient object-region tokens for memory attention. However, it does not reduce the storage footprint, as the original dense memory features are retained and further augmented with a sparse cache. Other saliency-based token merging~\cite{nguyen2026structsam, norouzi2024algm} and spatial pruning methods~\cite{liu2024surgical} primarily operate in the spatial domain~\cite{xiong2025efficient}, which can make them sensitive to video content. Therefore, training-free methods that directly reduce memory storage while preserving baseline performance remain largely underexplored~\cite{sun2026efficientvariantssegmentmodel}.

\paragraph{Tensor Decomposition.} 
Tensor decomposition has emerged as a robust solution for low-rank approximation and tensor compression~\cite{KoBa09}. Initial efforts in deep learning were primarily limited to using Singular Value Decomposition (SVD) to compress fully connected and convolutional layers~\cite{jaderberg2014speedingconvolutionalneuralnetworks}. More recent advancements, such as Generalized Kronecker Product Decomposition (GKPD), capture complex structural correlations in both linear and convolution layers~\cite{tai2016convolutionalneuralnetworkslowrank}. Moreover, PowerSGD~\cite{NEURIPS2019_d9fbed9d} leverages warm-started subspace iteration for gradient compression, demonstrating that reusing projection bases across iterations maintains accuracy while drastically reducing computational overhead. Similarly,~\citet{Nguyen2024ActivationMC,nguyen2025beyond}
explored activation map compression through low-rank decomposition,
demonstrating its effectiveness in alleviating activation memory
bottlenecks during backpropagation. Building on these advances, WASI~\cite{nguyen2026efficient}
further restricts training to a low-rank subspace, reducing memory consumption during backpropagation while improving inference efficiency in transformer models. In contrast to these approaches, which primarily focus on model parameters and training activations, we exploit the stable low-rank structure of streaming memory representations for efficient memory bank compression. Furthermore, we introduce a low-rank retrieval mechanism that avoids unnecessary reconstruction and computational overhead.
\section{Problem Statement}
\label{sec:problem_statement}

\paragraph{The Pipeline of SAM~2.}
The SAM2 image encoder $\text{E}_{\text{img}}(\cdot)$ adopts the Hiera architecture~\cite{hiera} as its backbone to hierarchically encode multi-scale image features. Defining the stages as $\{s_0, s_1, s_2\}$, the hierarchical feature set $\mathcal{F}_t$ for a given current frame $I_t$ is extracted as follows:
\begin{equation}
\mathcal{F}_t = \{F^{s_0}_t, F^{s_1}_t, F^{s_2}_t\} = \text{E}_{\text{img}}(I_t).
\end{equation}
Then $F^{s_2}_t$ is fused with the dense Memory Bank $\mathcal{M}_t$ (defined in Eq.~\eqref{eq:memory_bank}) within the memory attention module $\text{MA}(\cdot)$~\cite{ravi2024sam2} to obtain the memory-conditioned image feature $F_{M,t}$ as follows:
\begin{equation}
F_{M,t} = \text{MA}(F_t^{s_2}, \mathcal{M}_t) 
\label{eq:memory_attention}
\end{equation}
Here, $F^{s_2}_t$ acts as the query, while the dense Memory Bank $\mathcal{M}_t$  provides the keys and values.

Next, the mask decoder $\text{D}(\cdot)$ performs cross-attention between a prompt embedding $P_t$ and image features $F_{M,t}, F^{s_0}_t, F^{s_1}_t$ to obtain the mask prediction $O_t$ \cite{ravi2024sam2}:
\begin{equation}
O_t = \text{D}(F_{M,t}, F^{s_0}_t, F^{s_1}_t, P_t).
\end{equation}

Finally, $O_t$ and $F^{s_0}_t$ are fused and encoded as a new memory entry $M_t \in \mathbb{R}^{B \times C \times H \times W}$ in a first-in-first-out (FIFO) manner by the memory encoder $\text{E}_{\text{mem}}(\cdot)$:
\begin{equation}
M_{t} = \text{E}_{\text{mem}}(F^{s_0}_t, O_t),
\end{equation}
where $B$, $C$, $H$, and $W$ denote the batch size, number of channels, height, and width, respectively.

\paragraph{The Growing Size of the Memory Bank.} 
At time step $t$, the Memory Bank $\mathcal{M}_t$ comprises prompted frames $\mathcal{M}^p_t$ and a FIFO queue $\mathcal{M}^q_t$ of previous frames~\cite{ravi2024sam2}, formulated as:
\begin{equation}
    \mathcal{M}_t = \mathcal{M}^p_t \oplus \mathcal{M}^q_t = \mathcal{M}^p_t \oplus \{M_{t-N}, M_{t-N+1}, \dots, M_{t-1}\},
    \label{eq:memory_bank}
\end{equation}
where $N$ denotes the length of the queue and $\oplus$ represents concatenation along the temporal dimension.

Let $S$ denote the storage of a single memory tensor $M_t$. Given $N$ memory entries, the total storage cost is
\begin{equation}
    \mathcal{S}_{\mathrm{mem}} = N \cdot S = \mathcal{O}(N).
\end{equation}
Let $N_t$ denote the number of stored memory entries at time step $t$. As the number of stored frames increases with video length, \ie $N_t \propto t$, the cumulative Memory Bank size grows linearly (verified in Sec.\ref{sec:main_results}):
\begin{equation}
    \mathcal{S}_{\mathrm{mem}}(t) = N_t S = \mathcal{O}(tS).
\end{equation}
\section{Method}
\label{sec:method}

\begin{algorithm}[t]
\caption{MSI at iteration $t$}
\label{alg:msi}
\begin{algorithmic}[1]

\State \textbf{Input:} Memory tensor $M_{(t)}$, explained variance threshold $\varepsilon$, previous factors $\{\mathbf{U}_n^{(t-1)}\}$.

\State \textbf{Output:} Core $\mathcal{S}^{(t)}$, factors $\{\mathbf{U}_n^{(t)}\}$, and rank vector $\mathbf{r}=[r_B,r_C,r_H,r_W]$.

\If{$t = 0$}
    \State $\mathbf{r} \leftarrow$ determine ranks via $\varepsilon$ \Comment{see Eq.~\ref{rankfromvar}}
    \State $\mathcal{S}^{(t)}, \{\mathbf{U}_n^{(t)}\} \leftarrow \text{HOSVD}({M}_{(0)}, \mathbf{r})$ \Comment{see Eq.~\ref{eq:HOSVD}}
\Else
    \State $\mathcal{S} \leftarrow {M}_{(t)}$
    \For{$n = 1$ \textbf{to} $4$} \Comment{Modes: $B, C, H, W$}
        \State $\mathbf{A}_n \leftarrow \operatorname{unfold}(\mathcal{S}, n)$ \Comment{see Eq.~\ref{rankfromvar}}
        \State $\mathbf{V}_n \leftarrow \mathbf{A}_n^\top \mathbf{U}_n^{(t-1)}$ \Comment{Warm-start}
        \State $\mathbf{U}_n^{(t)} \leftarrow \operatorname{QR}(\mathbf{A}_n \mathbf{V}_n)$ \Comment{Subspace iteration}
        \State $\mathcal{S} \leftarrow \mathcal{S} \times_n \mathbf{U}_n^{(t)\top}$
    \EndFor
\EndIf

\State \Return $(\mathcal{S}^{(t)}, \{\mathbf{U}_n^{(t)}\}, \mathbf{r})$

\end{algorithmic}
\end{algorithm}

We first review low-rank tensor decomposition as the foundation for compact memory representation (Sec.~\ref{sec:tucker}). Next, we formulate our observation on the stable low-rank characteristics of memory tensors. We then introduce MSI to exploit this structure efficiently (Sec.~\ref{sec:msi}). Finally, we introduce a new low-rank retrieval mechanism (Sec.~\ref{sec:low-rank_retrieval}). An overview of our framework is illustrated in Fig.~\ref{fig:msi_pipeline}.

\subsection{Background} 
\label{sec:tucker}

\paragraph{SVD.} Given a matrix $\mathbf{A} \in \mathbb{R}^{M \times N}$, SVD factorizes it as
\begin{equation}
    \mathbf{A} = \mathbf{U} \mathbf{\Sigma} \mathbf{V}^T, \quad \mathbf{U} \in \mathbb{R}^{M \times M}, \mathbf{\Sigma} \in \mathbb{R}^{M \times N}, \mathbf{V} \in \mathbb{R}^{N \times N},
\end{equation}
where $\mathbf{\Sigma}$ is a rectangular diagonal matrix composed of $K$ singular values $\{s_i\}_{i=1}^{K}$, with $K$ being the rank of $\mathbf{A}$. Moreover, we can compute the amount of overall variance $\sigma^2_i$ explained by the $i^{\text{th}}$ pair of SVD vectors as $\sigma^2_i = s^2_i / \sum_j s^2_j$. 

Knowing that the singular values in $\mathbf{\Sigma}$ are ordered descending ($s_i \geq s_j, \forall i \leq j$) and given a desired explained variance threshold $\varepsilon \in (0, 1]$, we can find the truncation rank $r \in [1, K]$ such that 
\begin{equation}
    \sum_{i=1}^{r} \sigma^2_i \geq \varepsilon.
\label{rankfromvar}
\end{equation}
Then, $\mathbf{A}$ can be approximated by selecting only the first $r$ columns of $\mathbf{U}$ and $\mathbf{V}$, along with the first $r$ singular values from $\mathbf{\Sigma}$:
\begin{equation}
    \tilde{\mathbf{A}} = \mathbf{U}_{[:,:r]} \mathbf{\Sigma}_{[:r]} \mathbf{V}^T_{[:r,:]},
\end{equation}
\begin{equation}
\quad \mathbf{U}_{[:,:r]} \in \mathbb{R}^{M \times r}, \mathbf{\Sigma}_{[:r]} \in \mathbb{R}^{r \times r}, \mathbf{V}_{[:r,:]} \in \mathbb{R}^{N \times r}.
\end{equation}

\paragraph{HOSVD.} While SVD applies to matrices, flattening tensors into matrices can destroy internal structural correlations~\cite{ou2023lowrankoptimizationefficient}. To preserve this structure, we adopt Tucker decomposition~\cite{Tucker_1966}, which generalizes the matrix factorization above to higher-order tensors. For an $n^{\text{th}}$ order tensor $\mathcal{T} \in \mathbb{R}^{M_1 \times M_2 \times \dots \times M_n}$, the decomposition is expressed as:
\begin{equation}
    \mathcal{T} = \mathcal{S} \times_1 \mathbf{U}_1 \times_2 \mathbf{U}_2 \times_3 \dots \times_n \mathbf{U}_n,
\end{equation}
where $\mathcal{S} \in \mathbb{R}^{L_1 \times L_2 \times \dots \times L_n}$ is the core tensor (with $L_j$ the size of mode $j$ in the core) and $\mathbf{U}_j \in \mathbb{R}^{M_j \times L_j}$ represents the principal components over the $j^{\text{th}}$ mode. The $i$-mode product $\times_i$ between an $n^{\text{th}}$ order tensor $\mathcal{G} \in \mathbb{R}^{P_1 \times \dots \times P_n}$ and a matrix $\mathbf{B} \in \mathbb{R}^{Q \times P_i}$ is defined as:
\begin{equation}
    (\mathcal{G} \times_i \mathbf{B})_{p_1, \dots, p_{i-1}, q, p_{i+1}, \dots, p_n} = \sum_{p_i=1}^{P_i} g_{p_1, \dots, p_n}\, b_{q, p_i}.
\end{equation}

HOSVD is a special case of Tucker decomposition in which the factor matrices $\mathbf{U}_j$ are obtained by applying SVD to each mode-wise unfolding of the tensor, making them orthogonal~\cite{multiliner}. Given an explained variance threshold $\varepsilon$, the reconstructed tensor $\widetilde{\mathcal{T}}$ is formulated as:
\begin{equation}
    \widetilde{\mathcal{T}} = \hat{\mathcal{S}} \times_1 \mathbf{U}_{1_{[:,:r_1]}}\times_2 \dots \times_n \mathbf{U}_{n_{[:,:r_n]}} \approx \mathcal{T},
\label{eq:HOSVD}
\end{equation}
where $r_j$ is the truncation rank determined by $\varepsilon$ for mode $j$, and $\hat{\mathcal{S}} \in \mathbb{R}^{r_1 \times \dots \times r_n}$ is the truncated version of the core tensor $\mathcal{S}$.

\subsection{Memory Subspace Iteration}
\label{sec:msi}

\paragraph{Stable Low-rank Nature of Video Frames.}Prior work has shown that deep neural network features often lie in low-dimensional subspaces~\cite{li2018measuringintrinsicdimensionobjective,denton2014exploitinglinearstructureconvolutional}. We extend this observation to memory tensors in streaming video architectures. In many practical applications, such as smart cameras, robotics, etc, the input video stream is typically continuous and does not contain abrupt scene cuts. This motivates two observations:

\begin{itemize}
    \item \textit{Intrinsic Low-rank Structure.} Memory tensors $M_t$ reside in low-dimensional subspaces, implying that their semantic information can be effectively preserved by low-rank approximations.
    
    \item \textit{Temporal Subspace Stability.} Due to temporal coherence in continuous video frames~\cite{ravanbakhsh2024deepvideorepresentationlearning,AlShaykh2014Academic,deka2026inferencetimetemporalprobabilitysmoothing}, the subspace carrying the dominant information changes slowly over time. While the explicit pixel values evolve, the low-rank subspace explaining most of the variance can be considered stable over a sufficiently short temporal window.
\end{itemize}

\paragraph{Memory Subspace Iteration. } Based on these observations, we propose MSI, a training-free method that compresses each memory tensor $M_t \in \mathbb{R}^{B \times C \times H \times W}$. The intrinsic low-rank property naturally motivates leveraging tensor decomposition for memory compression. 

Naively, a direct candidate is HOSVD, as described in Sec.~\ref{sec:tucker}, which preserves the multi-dimensional structure across batch, channel, and spatial modes. However, applying HOSVD to every memory frame is extremely costly and impractical for real-world deployment. An ablation study demonstrates its computational complexity in Supp.Mat..

MSI leverages the temporal stability of low-rank subspaces. Instead of performing HOSVD estimation for each frame, MSI performs HOSVD once on the initial frame and uses subspace iteration~\cite{nguyen2026efficient} for subsequent frames. MSI therefore consists of two main steps:

\begin{figure}[t]
    \centering
    \includegraphics[width=1.0\linewidth]{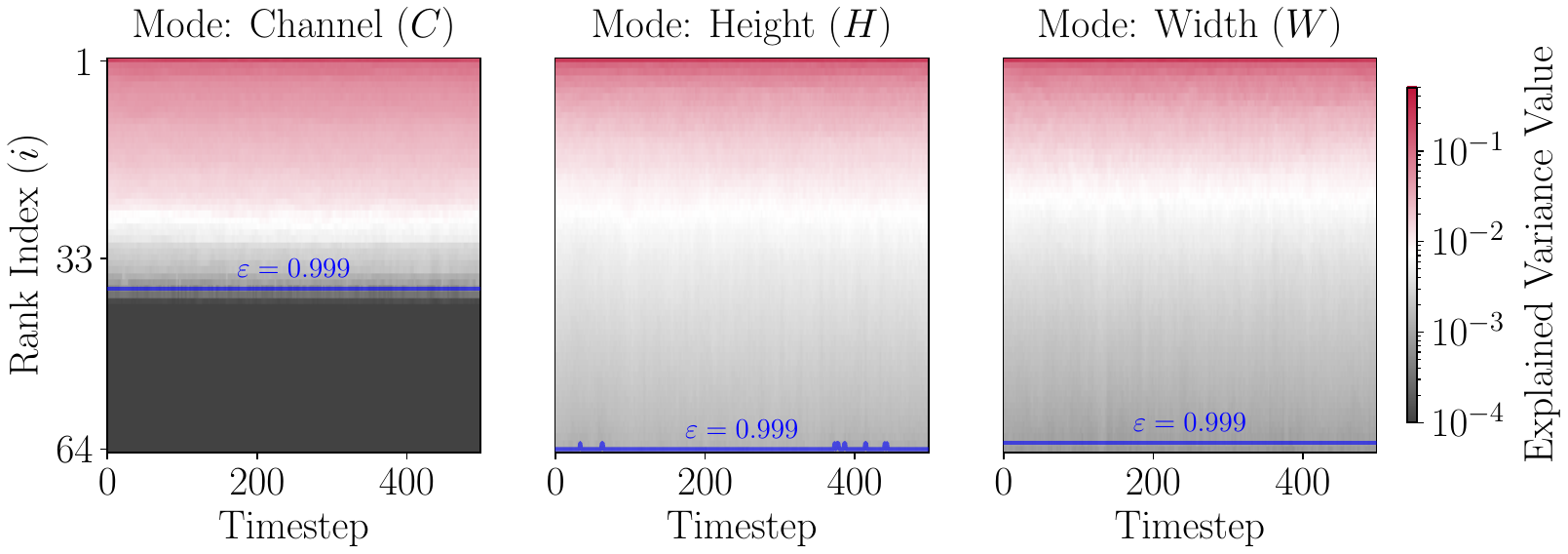}
    \vspace{-2mm}
    \caption{Singular value spectrum across 500 memory features from the SA-V test set on SAM2.1-B+, showing high channel-wise compressibility.}
    \label{fig:explained_variance_sav_v}
    \vspace{-3mm}
\end{figure}

\underline{Step 1:} MSI applies HOSVD once to the first memory frame to obtain the initial low-rank subspace. The truncation ranks are determined by an explained variance threshold $\varepsilon$, which controls information loss.

\underline{Step 2:} For subsequent frames, MSI uses the previous subspace as a warm start and applies Subspace Iteration to update the low-rank subspace efficiently.

The full procedure is described in Algorithm~\ref{alg:msi}.

\subsection{Low-Rank Memory Attention}
\label{sec:low-rank_retrieval}
To further leverage the low-rank structure, we design a new memory attention unit to operate directly on the low-rank components. This reduces the peak memory usage and computational cost. We formulate the new memory attention as:
\begin{equation}
F_{M,t} = \mathrm{MA}\left(F_t^{s_2},\, f_{\mathrm{MSI}}(\mathcal{M}_t)\right),
\label{eq:memory_attention_improved}
\end{equation}
where $\mathcal{M}_t$ is the low-rank Memory Bank obtained by storing each entry $M_t$ in its decomposed form, and $f_{\mathrm{MSI}}(\cdot)$ denotes a low-rank linear projection operator that reconstructs the keys and values for memory attention on the fly from the compressed factors (details in Supp.Mat.).

We first recall the standard scaled dot-product attention~\cite{vaswani2017attention}:
\begin{equation}
\mathrm{Attention}(\mathbf{Q}, \mathbf{K}, \mathbf{V})
=
\mathrm{Softmax}\!\left(\frac{\mathbf{Q}\mathbf{K}^\top}{\sqrt{d}}\right)\mathbf{V}.
\end{equation}

In SAM2, memory attention conditions current-frame features on past frames, predictions, and prompts using $L$ Transformer blocks, each consisting of self-attention, cross-attention over a Memory Bank, and an MLP. The query, key, and value are defined as:
\begin{align}
\mathbf{Q} = F^{s_2}_t \mathbf{W}_Q, \quad
\mathbf{K} = \mathbf{X}_K \mathbf{W}_K, \quad
\mathbf{V} = \mathbf{X}_V \mathbf{W}_V.
\label{eq:15}
\end{align}
\paragraph{Memory Construction.}

SAM2 forms memory tokens by concatenating flattened visual memory features and object pointers.

The flattened Memory Bank is:
\begin{equation}
\bar{\mathcal{M}}_{t} \in \mathbb{R}^{N_{\text{mem}} \times B \times C}, 
\quad N_{\text{mem}} = H \cdot W,
\end{equation}

and the full memory input is:
\begin{equation}
\mathbf{X}
=
\bar{\mathcal{M}}_{t} \oplus \mathbf{P}_{\text{obj}}
\in \mathbb{R}^{N \times B \times C},
\quad N = N_{\text{mem}} + N_{\text{ptr}}.
\end{equation}

where $\bar{\mathcal{M}}_{t}$ is obtained by flattening spatial feature maps from previous frames, and $\mathbf{P}_{\text{obj}}$ encodes object identity embeddings consistent across time. As more memory entries are retrieved, $\bar{\mathcal{M}}_{t}$ expands, directly leading to high VRAM usage.

We further construct positional embeddings as:
\begin{equation}
\mathbf{P}
=
\mathbf{Pos}_{\text{feat}} \oplus \mathbf{Pos}_{\text{obj}}
\in \mathbb{R}^{N \times B \times C}.
\end{equation}

The key and value inputs are defined as:
\begin{equation}
\mathbf{X}_K = \mathbf{X} + \mathbf{P}, \quad
\mathbf{X}_V = \mathbf{X}.
\end{equation}

Finally, projections are computed as:
\begin{equation}
\begin{aligned}
\mathbf{K}
&= \mathbf{X}_K \mathbf{W}_K, \\
\mathbf{V}
&= \mathbf{X}_V \mathbf{W}_V.
\end{aligned}
\label{K,V}
\end{equation}

where $\mathbf{W}_K$ and $\mathbf{W}_V$ are pretrained projection matrices inherited from SAM2.1, used without finetuning.

We avoid full $B \times C \times H \times W$ reconstruction and flattening, and instead project directly in the low-rank space before tokenization.

\begin{figure}
    \centering
    \includegraphics[width=0.92\linewidth]{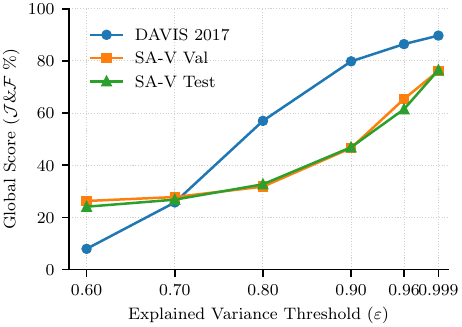}
   \caption{Increasing $\varepsilon$ consistently improves the average $\mathcal{J}\&\mathcal{F}$ score across DAVIS~2017, SA-V Val, and SA-V Test.}
    \label{fig:dynamicepsilon}
\end{figure}

\paragraph{MSI Factorization.}
With MSI, the memory tensor
$\mathcal{M}^{(t)} \in \mathbb{R}^{B \times C \times H \times W}$
is decomposed as:
\begin{equation}
\mathcal{M}^{(t)}
\approx
\mathcal{S}
\times_1 \mathbf{U}^{(B)}
\times_2 \mathbf{U}^{(C)}
\times_3 \mathbf{U}^{(H)}
\times_4 \mathbf{U}^{(W)} .
\label{eq:25}
\end{equation}
where $\mathcal{S} \in \mathbb{R}^{r_B \times r_C \times r_H \times r_W}$ is the core tensor.

The factor matrices are defined as \( \mathbf{U}^{(B)} \in \mathbb{R}^{B \times r_B} \), \( \mathbf{U}^{(C)} \in \mathbb{R}^{C \times r_C} \), \( \mathbf{U}^{(H)} \in \mathbb{R}^{H \times r_H} \), and \( \mathbf{U}^{(W)} \in \mathbb{R}^{W \times r_W} \), with \( r_B = B \), \( r_C \ll C \), \( r_H < H \), and \( r_W < W \).

We determine the rank vector $\mathbf{r} = [r_B, r_C, r_H, r_W]$ from explained variance threshold $\varepsilon$ in Eq.~\ref{rankfromvar}.

\paragraph{Low-rank Projection Fusion.}
In the standard memory projection, we denote Y as {K, V}, which is defined as:
\begin{equation}
\mathbf{Y}_{b,h,w,d}
=
\sum_{c=1}^{C}
\mathcal{M}^{(t)}_{b,c,h,w}\mathbf{W}_{c,d}.
\end{equation}

Using the channel-wise low-rank basis $\mathbf{U}^{(C)}$, the projection matrix is compressed as:
\begin{equation}
\tilde{\mathbf{W}} = (\mathbf{U}^{(C)})^\top \mathbf{W}, \quad
\tilde{\mathbf{W}}_{r_C,d}
=
\sum_{c=1}^{C}
\mathbf{U}^{(C)}_{c,r_C}\mathbf{W}_{c,d}.
\end{equation}

Substituting the MSI factorization yields:
\begin{align}
\mathcal{Z}_{br_C hw}
&=
\sum_{r_B,r_H,r_W}
\mathcal{S}_{r_B r_C r_H r_W}
U^{(B)}_{br_B}U^{(H)}_{hr_H}U^{(W)}_{wr_W},\\
\mathbf{Y}_{bhwd}
&=
\sum_{r_C}\mathcal{Z}_{br_C hw}\tilde{\mathbf{W}}_{r_Cd}.
\label{eq:27}
\end{align}

This formulation avoids explicit reconstruction of $\mathcal{M}^{(t)}$.
\section{Experiments}
\label{sec:experiments}
In this section, we evaluate MSI in terms of segmentation accuracy, memory efficiency, computational cost, and deployment efficiency. We first describe the experimental setup in Sec.~\ref{sec:setup}. Next, we conduct ablation studies on the low-rank structure, HOSVD comparison, and explained variance threshold in Sec.~\ref{sec:ablation}. Finally, Sec.~\ref{sec:main_results} presents results on standard benchmarks, Memory Bank compression, hardware efficiency, and edge deployment.

\subsection{Experimental Setup}
\label{sec:setup}
\begingroup
\renewcommand{\thefootnote}{}
\footnote{Additional implementation details are provided in Supp.Mat.}
\endgroup
We evaluate MSI across multiple backbones, including SAM~2.1 (\texttt{B+} and \texttt{L}), SAM3, and SAM2Long. We compare it against other efficient methods, including Efficient-SAM2 (ES2) and EdgeTAM, at two temporal sampling intervals ($\Delta t \in \{1, 5\}$). Tracking quality is evaluated using the average $\mathcal{J}\&\mathcal{F}$ score ($\mathcal{J}$ evaluates mask/region accuracy; $\mathcal{F}$ evaluates boundary accuracy) across seven standard VOS benchmarks: SA-V(T), DAVIS~2017, SA-V(V), MOSEv2, SeCVOS, LVOS, and YouTube-VOS (Table~\ref{tab:main}).

\begin{figure}[t]
    \centering
    \includegraphics[width=0.82\linewidth]{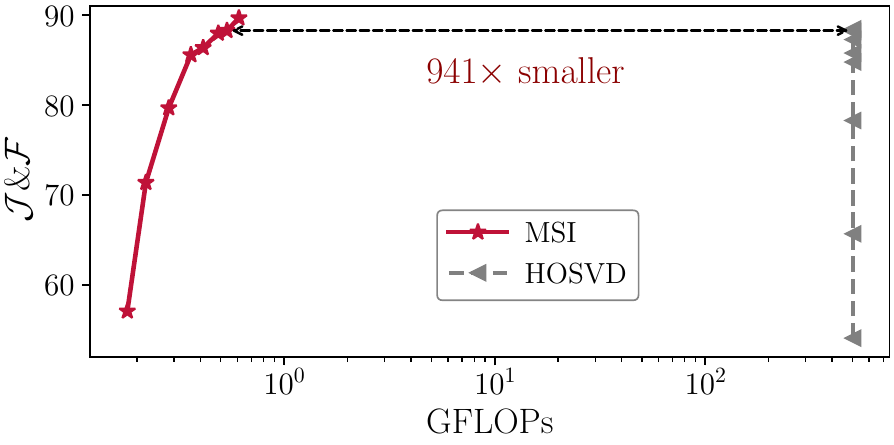}
    \caption{Performance ($\mathcal{J}\&\mathcal{F}$) and computational cost (GFLOPs) on SAM~2.1-B+ across $\varepsilon \in \mathcal{E}$ between MSI and HOSVD.}
    \label{fig:msivshosvd}
    \vspace{-3mm}
\end{figure}

\begin{table*}[t]
	\centering
	\caption{Quantitative tracking performance ($\mathcal{J}\&\mathcal{F}$) and peak GPU memory across 7 VOS benchmarks. Evaluation under temporal sampling intervals $\Delta t \in \{1, 5\}$. Peak GPU memory is measured on an NVIDIA A100 (40\,GB) with $M=7$. \textbf{Bold} and \underline{underline} denote the best and second-best results within each model subgroup, respectively. \colorbox{blue!8}{Blue-shaded rows} highlight our proposed MSI methods.}
	\label{tab:main}
	\setlength{\tabcolsep}{3.5pt}
	\renewcommand{\arraystretch}{1.2}
	\footnotesize
	\resizebox{\textwidth}{!}{%
		\begin{tabular}{ll ccccccc ccccccc c c}
			\toprule
			\multirow{2}{*}{\textbf{Backbone}}
			& \multirow{2}{*}{\textbf{Method}}
			& \multicolumn{7}{c}{\textbf{$\Delta t = 1$ ($\mathcal{J}\&\mathcal{F}\uparrow$)}}
			& \multicolumn{7}{c}{\textbf{$\Delta t = 5$ ($\mathcal{J}\&\mathcal{F}\uparrow$)}}
			& \multirow{2}{*}{\textbf{Avg} $\uparrow$}
			& \multirow{2}{*}{\textbf{Mem. (GB)} $\downarrow$} \\
			\cmidrule(lr){3-9} \cmidrule(lr){10-16}
			& & \textbf{SA-V(T)} & \textbf{DAVIS} & \textbf{SA-V(V)} & \textbf{MOSE} & \textbf{SeCVOS} & \textbf{LVOS} & \textbf{YTVOS}
			& \textbf{SA-V(T)} & \textbf{DAVIS} & \textbf{SA-V(V)} & \textbf{MOSE} & \textbf{SeCVOS} & \textbf{LVOS} & \textbf{YTVOS} & & \\
			\midrule
			
			% ==================== SAM3 ====================
			\multirow{2}{*}{\textbf{SAM3}}
			& Baseline
			& \textbf{84.0} & \textbf{91.5} & \textbf{82.6} & \textbf{59.8} & \textbf{73.7} & \textbf{88.6} & \textbf{89.6}
			& \textbf{81.1} & \textbf{91.6} & \textbf{82.2} & \textbf{53.1} & \textbf{75.2} & \textbf{87.7} & \textbf{89.0}
			& \textbf{80.7} & \underline{17.13} \\
			
			& \ourcell{\textbf{MSI}}
			& \ourcell{\underline{79.8}} & \ourcell{\underline{90.3}} & \ourcell{\underline{80.8}} & \ourcell{\underline{59.2}} & \ourcell{\underline{73.3}} & \ourcell{\textbf{88.6}} & \ourcell{\textbf{89.6}}
			& \ourcell{\underline{78.6}} & \ourcell{\underline{90.5}} & \ourcell{\underline{80.0}} & \ourcell{\underline{53.0}} & \ourcell{\underline{74.7}} & \ourcell{\underline{87.0}} & \ourcell{\underline{88.9}}
			& \ourcell{\underline{79.6}} & \ourcell{\textbf{16.62}} \\
			
			\midrule
			
			% ==================== SAM2.1 B+ ====================
			\multirow{6}{*}{\textbf{SAM2.1-B+}}
			& SAM2Long~\cite{ding2025sam2longenhancingsam2}
			& \textbf{79.3} & \textbf{90.1} & \textbf{80.9} & \underline{47.6} & \underline{61.4} & \textbf{84.2} & \underline{87.9}
			& \textbf{79.3} & \textbf{90.1} & \textbf{80.9} & \underline{47.6} & \underline{61.4} & \textbf{84.7} & \underline{87.9}
			& \textbf{76.0} & \underline{16.57} \\
			
			& \ourcell{\textbf{MSI+SAM2Long}}
			& \ourcell{\underline{78.7}} & \ourcell{\underline{90.0}} & \ourcell{\underline{79.8}} & \ourcell{\textbf{47.7}} & \ourcell{\textbf{61.5}} & \ourcell{\underline{83.1}} & \ourcell{\textbf{88.1}}
			& \ourcell{\underline{78.9}} & \ourcell{\underline{90.0}} & \ourcell{\underline{79.9}} & \ourcell{\textbf{47.7}} & \ourcell{\textbf{61.5}} & \ourcell{\underline{83.1}} & \ourcell{\textbf{88.1}}
			& \ourcell{\underline{75.6}} & \ourcell{\textbf{15.65}} \\
			\cmidrule(lr){2-18}
			
			& Baseline
			& \textbf{77.2} & \textbf{89.7} & \textbf{77.1} & \underline{43.3} & \underline{57.3} & \underline{82.2} & \textbf{88.2}
			& \underline{78.8} & \textbf{88.6} & \textbf{79.8} & \textbf{46.8} & \underline{58.7} & \textbf{84.0} & \textbf{88.0}
			& \textbf{74.3} & 13.66 \\
			
			& \ourcell{\textbf{MSI}}
			& \ourcell{\underline{76.6}} & \ourcell{\textbf{89.7}} & \ourcell{\underline{76.9}} & \ourcell{\textbf{44.1}} & \ourcell{\textbf{57.6}} & \ourcell{\textbf{82.7}} & \ourcell{\underline{87.6}}
			& \ourcell{\textbf{79.2}} & \ourcell{\underline{88.5}} & \ourcell{\underline{78.8}} & \ourcell{\underline{46.6}} & \ourcell{\textbf{59.1}} & \ourcell{\underline{83.9}} & \ourcell{\underline{87.3}}
			& \ourcell{\underline{74.2}} & \ourcell{\underline{13.36}} \\
			
			& ES2~\cite{zhang2026efficient}
			& 70.8 & \textbf{89.7} & 72.5 & 40.9 & 55.7 & 75.6 & 85.6
			& 72.8 & \textbf{88.6} & 74.2 & 42.6 & 57.8 & 76.8 & 85.7
			& 70.7 & 22.83 \\
			
			& EdgeTAM~\cite{zhou2025edgetam}
			& 70.9 & 86.4 & 71.4 & 43.1 & 53.8 & 79.3 & 85.9
			& 71.6 & 86.2 & 72.9 & 43.9 & 53.9 & 81.1 & 85.8
			& 70.4 & \textbf{12.47} \\
			
			\midrule
			
			% ==================== SAM2.1 L ====================
			\multirow{5}{*}{\textbf{SAM2.1-L}}
			& SAM2Long~\cite{ding2025sam2longenhancingsam2}
			& \textbf{80.0} & \textbf{89.8} & \textbf{82.7} & \textbf{50.6} & \textbf{62.0} & \textbf{84.8} & \textbf{88.8}
			& \textbf{80.0} & \textbf{89.8} & \underline{80.9} & \textbf{50.6} & \textbf{62.0} & \textbf{84.9} & \textbf{88.8}
			& \textbf{76.8} & \underline{17.90} \\
			
			& \ourcell{\textbf{MSI+SAM2Long}}
			& \ourcell{\underline{79.5}} & \ourcell{\underline{89.7}} & \ourcell{\underline{82.1}} & \ourcell{\underline{50.5}} & \ourcell{\underline{60.4}} & \ourcell{\underline{81.5}} & \ourcell{\textbf{88.8}}
			& \ourcell{\underline{79.7}} & \ourcell{\underline{89.7}} & \ourcell{\textbf{82.1}} & \ourcell{\underline{50.5}} & \ourcell{\underline{60.2}} & \ourcell{\underline{81.5}} & \ourcell{\textbf{88.8}}
			& \ourcell{\underline{76.1}} & \ourcell{\textbf{16.97}} \\
			\cmidrule(lr){2-18}
			
			& Baseline
			& \textbf{78.4} & \textbf{89.9} & \textbf{79.0} & \textbf{49.4} & \underline{58.2} & \textbf{84.2} & \textbf{88.8}
			& \textbf{80.3} & \underline{89.9} & \underline{80.2} & \textbf{50.8} & \textbf{59.3} & \textbf{86.7} & \textbf{88.7}
			& \textbf{76.0} & \underline{14.95} \\
			
			& \ourcell{\textbf{MSI}}
			& \ourcell{\underline{78.2}} & \ourcell{\underline{89.7}} & \ourcell{\underline{78.8}} & \ourcell{\underline{48.9}} & \ourcell{\textbf{58.5}} & \ourcell{\underline{84.1}} & \ourcell{\textbf{88.8}}
			& \ourcell{\underline{80.0}} & \ourcell{\textbf{90.1}} & \ourcell{\textbf{80.5}} & \ourcell{\underline{50.7}} & \ourcell{\textbf{59.3}} & \ourcell{\underline{86.5}} & \ourcell{\textbf{88.7}}
			& \ourcell{\underline{75.9}} & \ourcell{\textbf{14.63}} \\
			
			& ES2~\cite{zhang2026efficient}
			& 74.6 & \underline{89.7} & 76.5 & 43.8 & 55.1 & 78.6 & 86.9
			& 75.8 & \underline{89.9} & 78.6 & 45.6 & 56.1 & 79.7 & 86.9
			& 72.7 & 28.66 \\

			\bottomrule
		\end{tabular}%
	}
	% \vspace{-3mm}
\end{table*}

\begin{figure*}[t]
    \centering
    \includegraphics[width=1.0\linewidth]{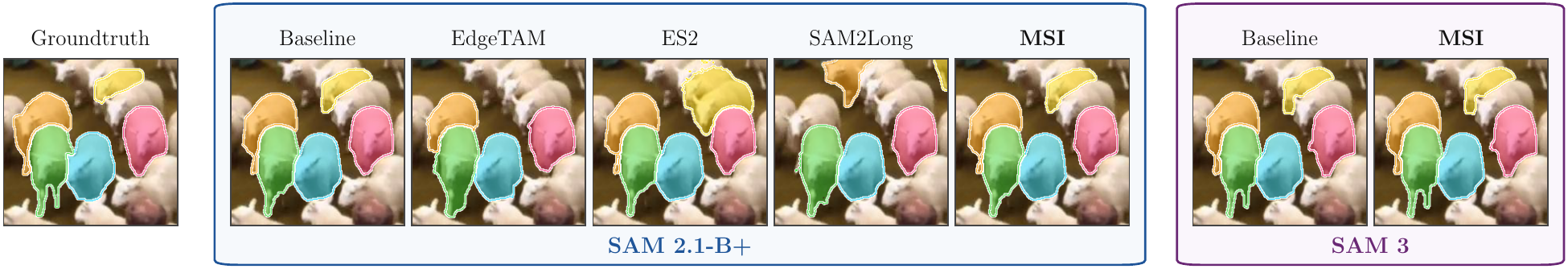}
    \vspace{-2mm}
    \caption{Close-up qualitative comparison across SAM2.1-B+ and SAM3 architectures. MSI preserves multi-object segmentation matching the ground truth, whereas prior efficient methods (EdgeTAM, ES2) exhibit mask drift.}
    \label{fig:main_qualitative_zoomin}
    \vspace{-3mm}
\end{figure*}

\subsection{Ablation Study}
\label{sec:ablation}

\paragraph{Stable Low-Rank Characteristics.}
We use SAM~2.1-B+ on the SA-V test set to extract the singular values of each memory-tensor mode across all video frames. As shown in Fig.~\ref{fig:explained_variance_sav_v}, the low-rank pattern remains relatively consistent over time, suggesting a temporally consistent subspace in the memory features. Additional experiments are provided in Supp.Mat.. Furthermore, the channel mode ($C$) requires only approximately $56\%$ of the full rank while the spatial modes ($H, W$) require nearly full rank to preserve 99.9\% of the energy. This indicates that redundancy mainly lies in the channel dimension, making it particularly amenable to compression. Together, these findings verify the stability of the low-rank structure in the Memory Bank, as introduced in Sec.~\ref{sec:msi}.

\paragraph{Explained Variance Threshold ($\varepsilon$).}
As shown in Fig.~\ref{fig:dynamicepsilon}, the tracking performance of SAM~2.1-B+ on DAVIS~2017 and SA-V improves monotonically with the explained variance threshold $\varepsilon$ and gradually approaches the baseline accuracy at $\varepsilon = 0.999$. Consequently, we adopt $\varepsilon = 0.999$ as our default benchmark setting.

\paragraph{MSI vs. HOSVD.}
We construct an HOSVD baseline that decomposes memory tensors independently at each inference step. We compare it with MSI on the DAVIS~2017-Val dataset across different values of $\varepsilon \in \mathcal{E}$ in Fig.~\ref{fig:msivshosvd}. Each marker on each curve corresponds to a different $\varepsilon$ value. MSI achieves comparable tracking performance to HOSVD at the same $\varepsilon$, while reducing the computational cost of memory decomposition by up to $941\times$. Additional analysis is provided in Supp.Mat.. These results demonstrate that, by exploiting the stable low-rank structure of memory representations, MSI can substantially reduce decomposition overhead without compromising segmentation quality relative to vanilla HOSVD.

\begin{figure}[t]
\centering
\includegraphics[width=1.0\linewidth]{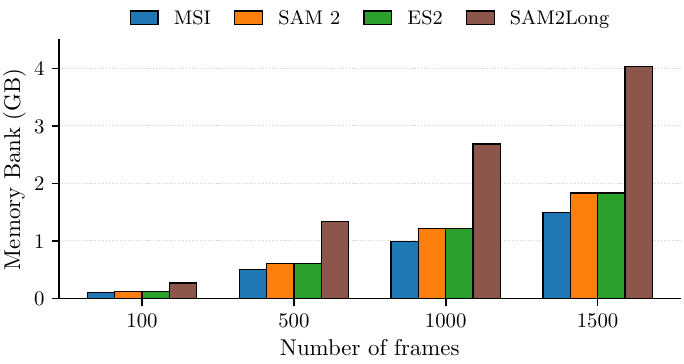}
\vspace{-2mm}
\caption{Memory bank size across different sequence lengths. Comparison of accumulated memory storage (GB) between MSI and baseline streaming architectures up to $1{,}500$ frames on SAM2.1-B+.} 
\label{fig:memory_bank_sam2_msi}
\vspace{-3mm}
\end{figure}

\subsection{Main Results}
\label{sec:main_results}

\begin{figure}[ht]
    \centering
    \begin{subfigure}{\linewidth}
        \centering
        \includegraphics[width=1.0\linewidth]{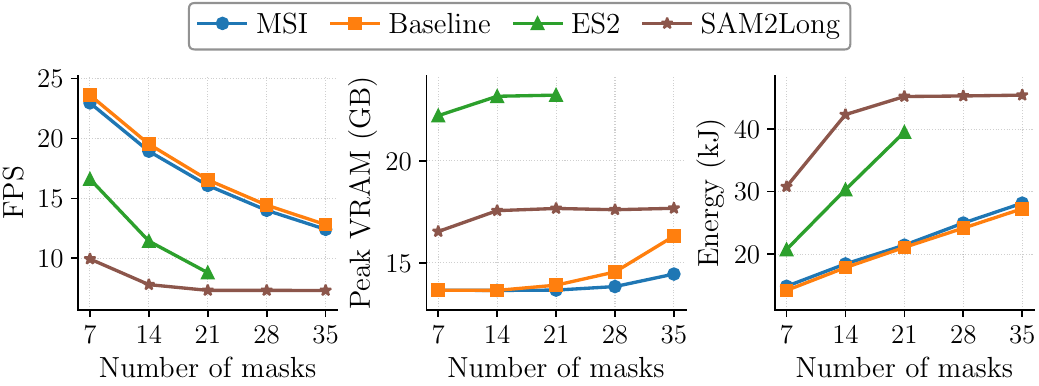}
        % \caption{NVIDIA RTX 3090 (24GB)}
        \label{fig:fps_3090_base}
    \end{subfigure}

    % \vspace{0.4cm}

    \begin{subfigure}{\linewidth}
        \centering
        \includegraphics[width=1.0\linewidth]{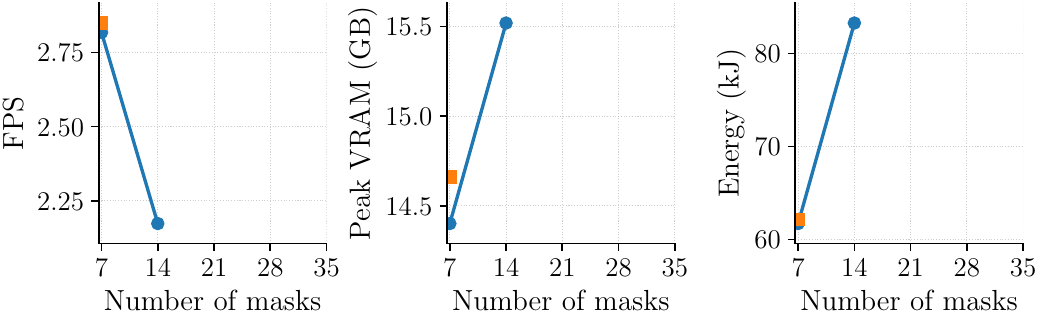}
        % \caption{NVIDIA Tesla P100 (16GB)}
        \label{fig:fps_p100_base}
    \end{subfigure}

  \caption{Comparison of inference throughput (FPS), peak reserved VRAM (GB), and total energy (kJ) on SAM2.1-B+. The experiments are conducted on (\textit{top}) NVIDIA RTX 3090 , and (\textit{bottom}) NVIDIA Tesla P100.} 
\vspace{-4mm}
    \label{fig:memory_base}
\end{figure}

\begin{figure*}[t]
    \centering
    \includegraphics[width=1.0\linewidth]{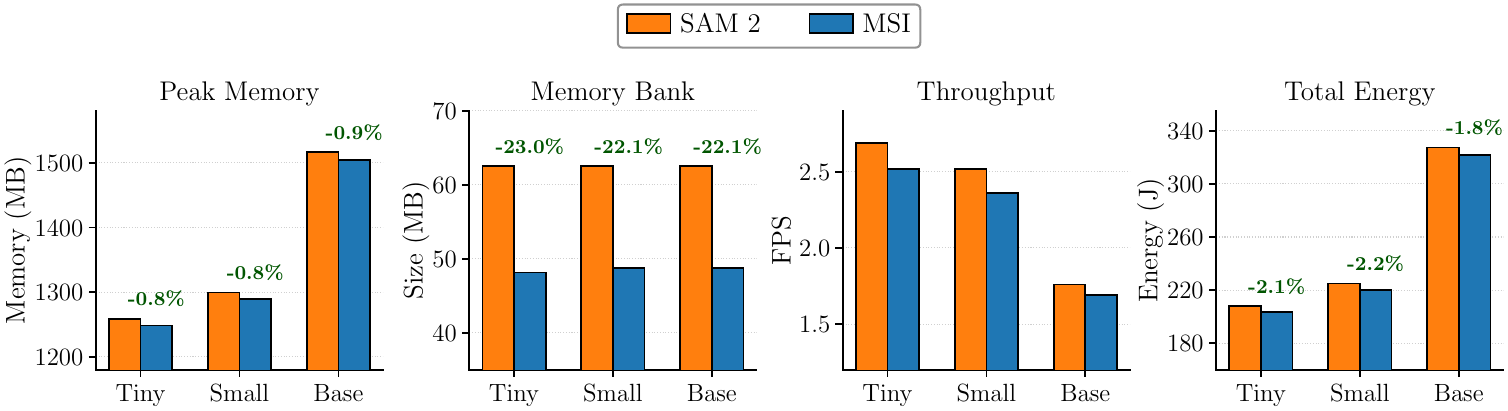}
    \vspace{-2mm}
    \caption{Edge benchmarking on NVIDIA Jetson Orin with 8\,GB of VRAM across model scales on DAVIS~2017 (50 frames, $M=7$).}
    \label{fig:jetson}
    % \vspace{-3mm}
\end{figure*}

Table~\ref{tab:main} reports results across seven diverse benchmarks and two sampling intervals ($\Delta t=1, 5$) on an NVIDIA A100 GPU with 40\,GB of memory. As shown in Fig.~\ref{fig:main_qualitative_zoomin}, MSI preserves sharp object boundaries under dense occlusions. Additional visual comparisons are provided in Supp.Mat..

Overall, MSI consistently reduces peak memory consumption while maintaining comparable tracking performance across all evaluated architectures. On SAM~2.1-L and SAM~2.1-B+, MSI achieves nearly identical average $\mathcal{J}\&\mathcal{F}$ scores to the corresponding baselines, with a decrease of only $0.1$ points for each model. For the long-video architecture ${\text{SAM2Long}}$, MSI incurs only a $0.4$-point decrease in average $\mathcal{J}\&\mathcal{F}$ while reducing peak memory consumption by up to $5.6\%$. Similarly, on ${\text{SAM3}}$, MSI remains within $1.1$ points of the baseline average $\mathcal{J}\&\mathcal{F}$ score while achieving a $3.0\%$ reduction in peak memory.

MSI is a completely \textit{training-free} method. Notably, $\text{MSI}$ significantly outperforms EdgeTAM by $3.8$ average $\mathcal{J}\&\mathcal{F}$ points, despite requiring no task-specific fine-tuning. Furthermore, MSI surpasses ES2 by $3.5$ average $\mathcal{J}\&\mathcal{F}$ points while consuming up to $49.0\%$ less peak memory, avoiding the additional sparse-memory overhead introduced by ES2.

\paragraph{Memory Bank Compression.} 
As shown in Fig.~\ref{fig:memory_bank_sam2_msi}, MSI consistently reduces the accumulated Memory Bank footprint across extended sequence lengths ($N \in \{100, \dots, 1{,}500\}$ frames). Specifically, at $N=1{,}500$ frames, MSI reduces Memory Bank storage by approximately \textbf{$18.4\%$} compared to standard SAM2. Since each incoming memory entry is stored in its compressed representation, these savings accumulate throughout the sequence rather than being limited to a single retrieval step. This demonstrates that the channel-wise redundancy in Memory Banks can be effectively compressed. In contrast, heavy architectures such as SAM2Long and SAM3 accumulate up to $4.03$\,GB and $4.64$\,GB, respectively. Further analysis is provided in Supp.Mat..

\paragraph{Hardware Efficiency.}

Fig.~\ref{fig:memory_base} reports inference throughput (FPS), peak reserved VRAM (GB), and total energy consumption (kJ) across methods on an NVIDIA RTX~3090 (24 GB) and Tesla~P100 (16 GB). We vary the number of retrieval entries of SAM~2.1-B+ as $M \in \{7,\dots,35\}$ to evaluate scaling with memory depth. Additional experimental details and results are provided in Supp.Mat..

In terms of inference throughput, MSI closely matches vanilla SAM~2.1-B+ on both GPUs, with less than a $3\%$ reduction across retrieval depths. In contrast, Efficient-SAM2 runs approximately $30\%$ slower than both the baseline and MSI, while SAM2Long remains below $10$ FPS. This suggests that the overhead of dual dense-sparse memory management and long-term windowed attention can offset their intended computational savings during on-device inference.

In terms of memory usage, MSI maintains a bounded footprint as $M$ increases. On the RTX~3090, its peak reserved VRAM remains below $70\%$ of the physical GPU capacity even at $M=35$. By contrast, Efficient-SAM2 uses over $35\%$ more peak VRAM than MSI at $M=21$ and triggers Out-Of-Memory (OOM) errors at $M \ge 28$ due to memory-bank accumulation. The memory advantage is more pronounced on the 16 GB Tesla~P100, where vanilla SAM2.1-B+ encounters OOM at $M \ge 14$, whereas MSI remains operational at $M=14$ while using less than $90\%$ of the available VRAM.

Finally, MSI maintains relatively stable energy consumption as memory depth increases. On the RTX~3090, its energy usage remains close to that of the vanilla baseline, whereas SAM2Long incurs higher cumulative energy consumption due to its long-term windowing mechanism. Overall, MSI improves memory scalability while preserving near-baseline throughput and energy efficiency across both GPUs, demonstrating that these memory savings incur negligible overhead.

\paragraph{Edge Deployment.} We evaluate MSI using the SAM2.1-B+, Small, and Tiny models and their corresponding baselines on an NVIDIA Jetson Orin-8 GB using the DAVIS~2017-Val benchmark. As shown in Fig.~\ref{fig:jetson}, MSI maintains tracking performance comparable to the baseline while reducing the memory-bank footprint by up to $23.0\%$. In addition, MSI consistently lowers overall energy consumption by up to $3\%$ across video sequences, highlighting its potential for more sustainable edge deployment. These savings can be further increased by decreasing $\varepsilon$, at the cost of a corresponding degradation in tracking performance.

\section{Conclusion}
\label{sec:conclusion}
In this work, we presented Memory Subspace Iteration (MSI), a training-free method that mitigates memory growth in streaming video segmentation. By leveraging the temporal stability of low-rank memory representations, MSI substantially reduces memory footprint and energy consumption with negligible accuracy degradation. Ultimately, MSI helps bridge the gap between large-scale video foundation models and their deployment on resource-constrained devices. We hope these findings will provide a foundation for improving resource efficiency in broader multimodal systems, including world models.

\newpage
\section*{Acknowledgments} This work benefited from Hi! PARIS and State funding managed by the French National Research Agency (ANR) under the France 2030 program, reference ANR-23-IACL-0005. This work was also supported by the Italian MUR through the FIS 2 project FIS-2023-03417 (DD n. 23314 10-12-2024, CUP D53C25002040001).
{
    \small
    \bibliographystyle{ieeenat_fullname}
    \bibliography{main}
}
\appendix
\newpage
\appendix
\section{Additional Theoretical Details}
\label{app:theory}

In this section, we analyze the computational overhead of each compression method, the speedup in efficient fusion-retrieval computation, and the memory required to store compressed memory tensors.

\subsection{Memory Efficiency}
\label{sec:overhead}
For a single memory entry $M_t \in \mathbb{R}^{B \times C \times H \times W}$, let $\mathcal{D}=(\mathcal{D}_1, \mathcal{D}_2, \mathcal{D}_3, \mathcal{D}_4)=(B,C,H,W)$ denote the original tensor dimensions and $\mathbf{r}=(r_B,r_C,r_H,r_W)$ denote the truncation ranks determined by the explained variance threshold $\varepsilon$. Instead of storing the dense tensor with $\mathcal{O}_\text{space}\left(\prod_{m=1}^{4} \mathcal{D}_m\right)$ elements per entry, MSI stores only the compact low-rank representations, consisting of the core tensor and the mode-wise factor matrices. The resulting compression ratio is:
\begin{equation}
    \rho(\mathbf{r})
    =
    \frac{\prod_{m=1}^{4} \mathcal{D}_m}
    {    \prod_{m=1}^{4} r_m
    +
    \sum_{m=1}^{4} \mathcal{D}_m r_m},
    \quad r_m \ll \mathcal{D}_m.
    \label{eq:memory_compression}
\end{equation}

\begin{figure}[t]
    \centering
    \includegraphics[width=1.0\linewidth]{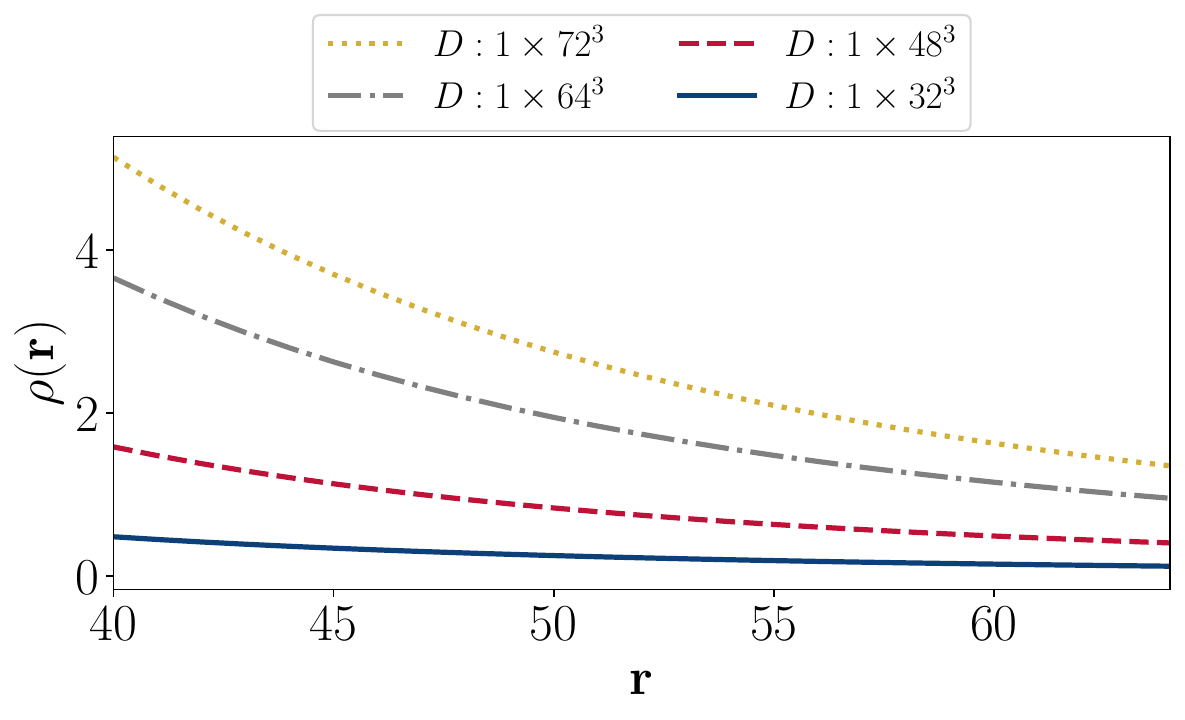}
    \vspace{-2mm}
    \caption{\textbf{Theoretical memory reduction ratio $\rho(\mathbf{r})$.} Compression scaling of MSI across truncation ranks $\mathbf{r}$ and tensor dimensions $\mathcal{D}$ (Eq.~\eqref{eq:memory_compression}).}
    \label{fig:fig_compression_ratio}
    \vspace{-3mm}
\end{figure}

Fig.~\ref{fig:fig_compression_ratio} analyzes the theoretical compression ratio $\rho(\mathbf{r})$ of MSI under different ranks $\mathbf{r}$ and tensor sizes $\mathcal{D}$. The results show that MSI becomes more effective as the memory tensor grows larger or the retained rank becomes smaller. This suggests that MSI is especially beneficial for longer or higher-resolution videos, where the memory bank can grow substantially and memory-efficient representations are critical for resource-constrained deployment~\cite{zhou2025edgetam}.

\subsection{Decomposition Complexity}
\label{ap:complexity}

\begin{figure}[t]
	\centering
	\includegraphics[width=1.0\linewidth]{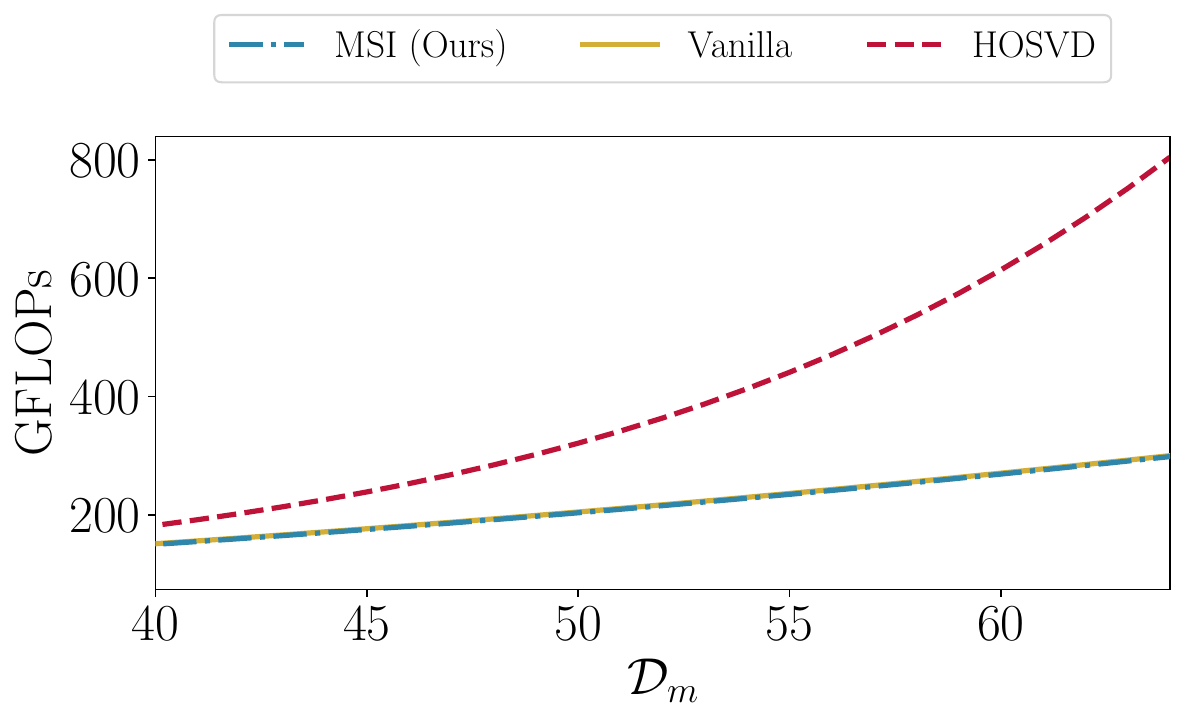}
	\vspace{-2mm}
	\caption{\textbf{Complexity comparison.} Theoretical GFLOPs of MSI, HOSVD, and vanilla baseline across tensor dimensions $\mathcal{D}$.}
	\label{fig:gflop}
	\vspace{-3mm}
\end{figure}

Fig.~\ref{fig:gflop} further compares the theoretical GFLOPs of MSI, HOSVD, and the vanilla baseline under different tensor sizes $\mathcal{D}$. As $\mathcal{D}$ increases, the cost of repeatedly applying HOSVD grows rapidly. In contrast, MSI remains close to the vanilla baseline, showing that it can provide stronger memory compression for larger videos without introducing substantial additional computational cost.

Using MSI to decompose the tensor is equivalent to performing Subspace Iteration on each of its modes~\cite{nguyen2026efficient}. Consider a memory tensor $\mathcal{M}_i \in \mathbb{R}^{D_{i,b} \times D_{i,c} \times D_{i,h} \times D_{i,w}}$. For mode $m \in \{b,c,h,w\}$, let $\mathcal{M}_{i,(m)} \in \mathbb{R}^{d \times d'}$ denote its unfolded matrix representation, where $d = D_{i,m}$ is the size along mode $m$, and $d' = \prod_{n \neq m} D_{i,n}$ is the product of all remaining dimensions.\\
Decomposing the unfolded tensor $\mathcal{M}_{i,(m)}$ via subspace iteration with rank $r_{i,m}$ requires
$\Theta(2 d_m P_m r_{i,m} + r_{i,m}^3)$ FLOPs. Therefore, the total computational cost of MSI across all four modes as is:
\begin{equation}
    \mathcal{O}_{\text{MSI}} =
    \sum_{m \in \{B,C,H,W\}}
    \left(
    2 d_m P_m r_{i,m} + r_{i,m}^3
    \right),
    \label{eq:msi_overhead}
\end{equation}
where $d_m$ denotes the size of mode $m$, and $P_m = \prod_{n \neq m} d_n$ is the product of all remaining dimensions.\\
For comparison, as shown in Nguyen et al.~\cite{Nguyen2024ActivationMC}, performing HOSVD on the same tensor $\mathcal{M}_i \in \mathbb{R}^{B \times C \times H \times W}$ incurs the following computational cost:
\begin{equation}
    \mathcal{O}_{\text{HOSVD}_\varepsilon}
    =
    \sum_{m \in \{B,C,H,W\}}
    \max(d_m, P_m)^2 \cdot \min(d_m, P_m),
    \label{eq:hosvd_overhead}
\end{equation}
Overall, MSI has lower decomposition complexity than HOSVD and scales more favorably with increasing tensor size, leading to reduced computational cost under identical input dimensions.

\paragraph{Decomposition FLOPs.}
We evaluate memory decomposition complexity on SAM~2.1-B+ ($M=7$ frames of shape $1 \times 64 \times 64 \times 64$). Naive HOSVD independently applies truncated SVD at each frame without warm start, accumulating $503.59$\,GFLOPs across the $M=7$ frames (Eq.~\eqref{eq:hosvd_overhead}). In contrast, MSI fixes rank once at Frame~0 and executes warm-started subspace iteration with factor matrix reuse (Eq.~\eqref{eq:msi_overhead}). At $\varepsilon \approx 0.945$ ($\mathcal{J}\&\mathcal{F} \approx 88.5\%$, rank $\mathbf{r}=[1, 22, 37, 40]$), MSI requires only $0.535$\,GFLOPs across 7 frames. This achieves a $\mathbf{941\times}$ FLOPs reduction ($503.59 / 0.535$), while delivering a $6.72\times$ wall-clock speedup ($1.21$\,ms vs.\ $8.12$\,ms) on an NVIDIA A100.

\subsection{Retrieval Framework}
\label{app:Retrieval}

We first recall the standard scaled dot-product attention~\cite{vaswani2017attention}:
\begin{equation}
\mathrm{Attention}(\mathbf{Q}, \mathbf{K}, \mathbf{V})
=
\mathrm{Softmax}\!\left(
\frac{\mathbf{Q}\mathbf{K}^\top}{\sqrt{d}}
\right)\mathbf{V}.
\end{equation}

In SAM~2.1, memory attention conditions current-frame features on past frames, predictions, and prompts using $L$ Transformer blocks, each consisting of self-attention, cross-attention over a memory bank, and an MLP. The query, key, and value are defined as:
\begin{align}
\mathbf{Q} &= F^{s_2}_t \mathbf{W}_Q, \nonumber\\
\mathbf{K} &= \mathbf{X}_K \mathbf{W}_K, \nonumber\\
\mathbf{V} &= \mathbf{X}_V \mathbf{W}_V.
\label{eq:15_ap}
\end{align}

\paragraph{Memory Construction.}

SAM~2.1 forms memory tokens by concatenating flattened visual memory features and object pointers.

The flattened memory bank is:
\begin{equation}
\bar{\mathcal{M}}_{t}
\in \mathbb{R}^{N_{\text{mem}} \times B \times C},
\qquad
N_{\text{mem}} = H \cdot W.
\end{equation}

and the full memory input is:
\begin{equation}
\mathbf{X}
=
\bar{\mathcal{M}}_{t} \oplus \mathbf{P}_{\text{obj}}
\in \mathbb{R}^{N \times B \times C},
\qquad
N = N_{\text{mem}} + N_{\text{ptr}}.
\end{equation}

where $\bar{\mathcal{M}}_{t}$ is obtained by flattening spatial feature maps from previous frames, and $\mathbf{P}_{\text{obj}}$ encodes object identity embeddings consistent across time.

We further construct positional embeddings as:
\begin{equation}
\mathbf{P}
=
\mathbf{Pos}_{\text{feat}} \oplus \mathbf{Pos}_{\text{obj}}
\in \mathbb{R}^{N \times B \times C}.
\end{equation}

The key and value inputs are defined as:
\begin{equation}
\mathbf{X}_K = \mathbf{X} + \mathbf{P},
\qquad
\mathbf{X}_V = \mathbf{X}.
\end{equation}

Finally, projections are computed as:
\begin{equation}
\begin{aligned}
\mathbf{K}
&= \mathbf{X}_K \mathbf{W}_K,\\
\mathbf{V}
&= \mathbf{X}_V \mathbf{W}_V.
\end{aligned}
\label{K_ap,V_ap}
\end{equation}

where $\mathbf{W}_K$ and $\mathbf{W}_V$ are pretrained projection matrices inherited from SAM~2.1, used without finetuning.

We avoid full $B \times C \times H \times W$ reconstruction and flattening, and instead perform projection directly in the low-rank space before tokenization.

\textbf{MSI Factorization.}
With MSI, the memory tensor
$\mathcal{M}^{(t)} \in \mathbb{R}^{B \times C \times H \times W}$
is decomposed as:
\begin{equation}
\begin{aligned}
\mathcal{M}^{(t)}_{b,c,h,w}
\approx
\sum_{r_B,r_C,r_H,r_W}
&
\mathcal{S}_{r_B,r_C,r_H,r_W}
\mathbf{U}^{(B)}_{b,r_B}
\mathbf{U}^{(C)}_{c,r_C}\\
&\cdot
\mathbf{U}^{(H)}_{h,r_H}
\mathbf{U}^{(W)}_{w,r_W},
\end{aligned}
\label{eq:25_ap}
\end{equation}
where $\mathcal{S} \in \mathbb{R}^{r_B \times r_C \times r_H \times r_W}$ is the core tensor.

The factor matrices are defined as
\( \mathbf{U}^{(B)} \in \mathbb{R}^{B \times r_B} \),
\( \mathbf{U}^{(C)} \in \mathbb{R}^{C \times r_C} \),
\( \mathbf{U}^{(H)} \in \mathbb{R}^{H \times r_H} \), and
\( \mathbf{U}^{(W)} \in \mathbb{R}^{W \times r_W} \), with
\( r_B = B \), \( r_C \ll C \), \( r_H < H \), and \( r_W < W \).

We determine the rank vector $\mathbf{r} = [r_B, r_C, r_H, r_W]$ from explained variance threshold $\varepsilon$.

Substituting Eq.~\ref{eq:25_ap} into Eq.~\ref{K_ap,V_ap}, we obtain:
\begin{equation}
\begin{aligned}
\mathbf{K}
=
\Bigg(
\sum_{r_B,r_C,r_H,r_W}
&
\mathcal{S}_{r_B,r_C,r_H,r_W}
\mathbf{U}^{(B)}_{:,r_B}
\mathbf{U}^{(C)}_{:,r_C}\\
&
\cdot
\mathbf{U}^{(H)}_{:,r_H}
\mathbf{U}^{(W)}_{:,r_W}
\Bigg)\mathbf{W}_K.
\end{aligned}
\end{equation}

\begin{equation}
\begin{aligned}
\mathbf{V}
=
\Bigg(
\sum_{r_B,r_C,r_H,r_W}
&
\mathcal{S}_{r_B,r_C,r_H,r_W}
\mathbf{U}^{(B)}_{:,r_B}
\mathbf{U}^{(C)}_{:,r_C}\\
&
\cdot
\mathbf{U}^{(H)}_{:,r_H}
\mathbf{U}^{(W)}_{:,r_W}
\Bigg)\mathbf{W}_V.
\end{aligned}
\end{equation}

\textbf{Low-rank Projection Fusion.}
The standard memory projection, we denoted Y $\in$ \{K, V\} is defined as:
\begin{equation}
\mathbf{Y}_{b,h,w,d}
=
\sum_{c=1}^{C}
\mathcal{M}^{(t)}_{b,c,h,w}
\mathbf{W}_{c,d}.
\end{equation}

Using the channel-wise low-rank basis $\mathbf{U}^{(C)}$, the projection matrix is compressed as:
\begin{align}
\tilde{\mathbf{W}}
&=
(\mathbf{U}^{(C)})^\top \mathbf{W},
\nonumber\\
\tilde{\mathbf{W}}_{r_C,d}
&=
\sum_{c=1}^{C}
\mathbf{U}^{(C)}_{c,r_C}
\mathbf{W}_{c,d}.
\end{align}

Substituting the MSI factorization yields:
\begin{align}
\mathcal{Z}_{b,r_C,h,w}
&=
\sum_{r_B,r_H,r_W}
\mathcal{S}_{r_B,r_C,r_H,r_W}
\mathbf{U}_{B,b,r_B}
\mathbf{U}_{H,h,r_H}
\mathbf{U}_{W,w,r_W},
\nonumber\\
\mathbf{Y}_{b,h,w,d}
&=
\sum_{r_C}
\mathcal{Z}_{b,r_C,h,w}
\tilde{\mathbf{W}}_{r_C,d}.
\label{eq:27_ap}
\end{align}

Therefore, $f_{\mathrm{MSI}}(\cdot)$ is defined by the procedure described in Eq.~\ref{eq:15_ap} to Eq.~\ref{eq:27_ap}.

This formulation avoids explicit reconstruction of $\mathcal{M}^{(t)}$, reducing the complexity from
$\mathcal{O}(BCHW)$ to
$\mathcal{O}(Br_B r_C r_H r_W + r_C HW)$,
and significantly reduces FLOPs in memory projection.

\subsection{Low-rank structure of memory tensors}
\label{apendix:qualitative_low_rank}

\begin{figure}[t]
    \centering

    % Row 1
    \begin{subfigure}{\linewidth}
        \centering
        \includegraphics[width=1.0\linewidth]{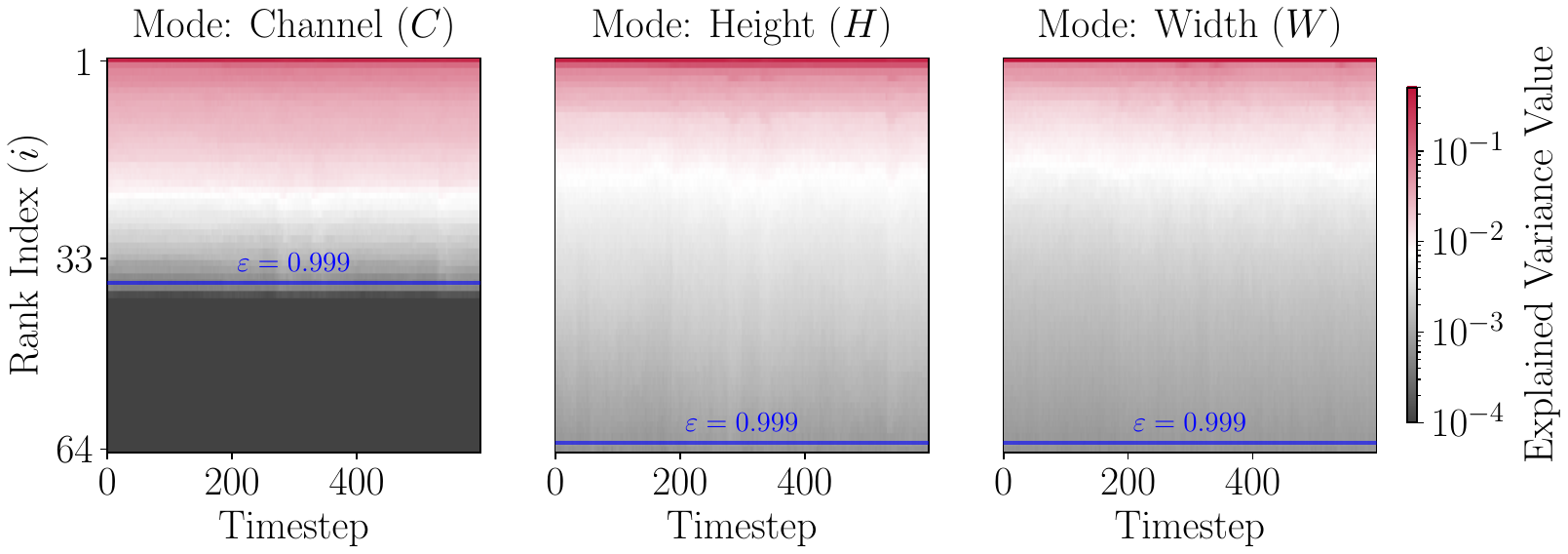}
        \vspace{-2mm}
        \caption{SA-V Val set}
    \end{subfigure}

    \vspace{1mm}

    % Row 2
    \begin{subfigure}{\linewidth}
        \centering
        \includegraphics[width=1.0\linewidth]{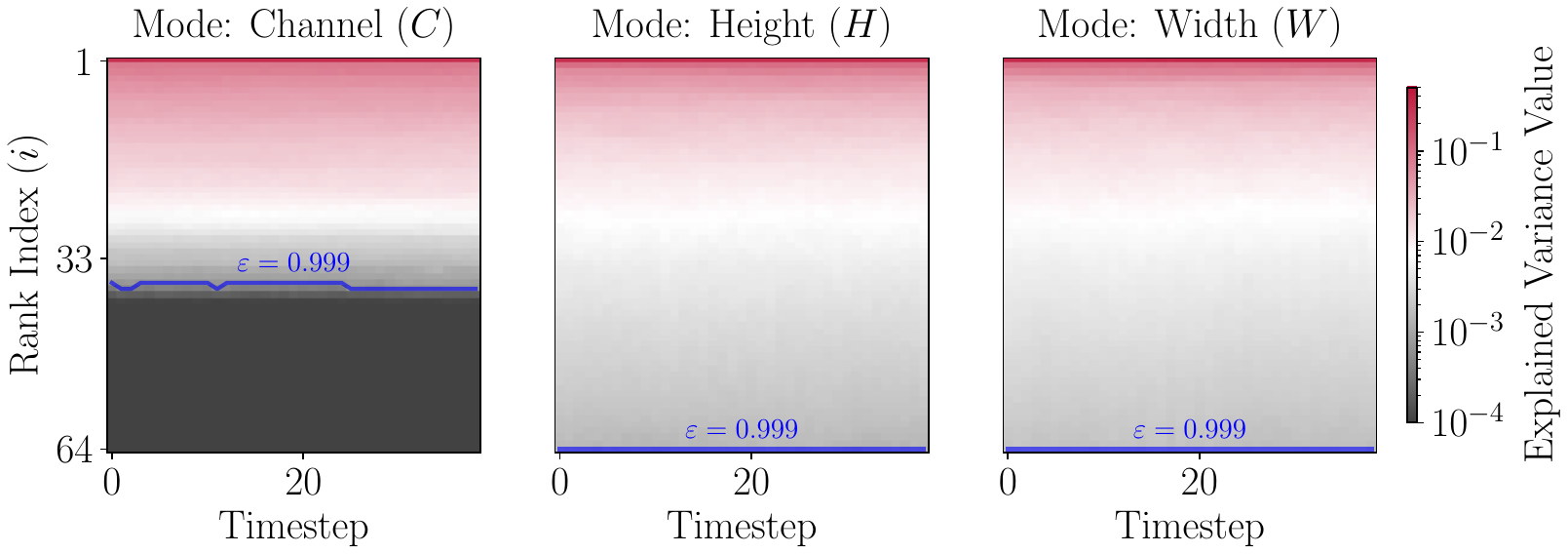}
        \vspace{-2mm}
        \caption{DAVIS 2017}
    \end{subfigure}
    \vspace{-2mm}
    \caption{\textbf{Explained variance spectrum of SAM~2.1.} Consistent spectral decay across SA-V Val (top) and DAVIS 2017 (bottom).}
    \label{fig:explained_variance_all}
    \vspace{-3mm}
\end{figure}

We evaluate the spectral properties of SAM~2.1-B+ memory tensors on DAVIS 2017~\cite{pont20172017} and SA-V~\cite{ravi2024sam2} (Test and Validation sets) in Fig.~\ref{fig:explained_variance_all}

\paragraph{Conclusion.}
Across both DAVIS 2017 and SA-V datasets, we consistently observe a stable low-rank structure in SAM~2.1 memory tensors across video content, duration, or temporal dynamics. The spectral decay patterns remain highly consistent across sequences, indicating that memory redundancy is an intrinsic nature of model architecture. This observation provides strong empirical justification for our MSI-based low-rank formulation.

\section{Additional Experimental Details}
\label{app:exp_details}

\subsection{Baseline}
\label{app:baseline}

\paragraph{SAM2~\cite{ravi2024sam2}.}
SAM2 is a foundation model for promptable image and video segmentation that maintains a memory bank of previously processed frames to support temporal propagation. During video inference, it retrieves relevant memory features through memory attention, enabling consistent object tracking across frames. In our experiments, we use the official SAM~2.1 checkpoints as the primary baseline. The default memory tensor has shape $(B, C, H, W) = (B, 64, 64, 64)$.

\paragraph{SAM3~\cite{carion2026sam3}.}
SAM3 extends SAM-based segmentation to support more general visual concepts and video understanding. It incorporates a memory mechanism to propagate object information over time while leveraging stronger visual representations. We evaluate MSI on SAM3 to examine whether the proposed memory compression framework generalizes to newer and larger video segmentation architectures. The default memory tensor has shape $(B, C, H, W) = (B, 72, 72, 72)$.

\subsection{Datasets}
\label{app:dataset}

\paragraph{SA-V~\cite{ravi2024sam2}.} The SA-V dataset contains 50.9K video clips and 642.6K masklets for video object segmentation. Its validation set contains 293 masklets across 155 videos, while the test set comprises 278 masklets from 150 videos. It encompasses challenging scenarios such as occlusions, small objects, and target reappearance, serving as a large-scale benchmark for modern VOS models. In our experiments, we evaluate on the first 100 videos of the validation (SA-V(V)) and test (SA-V(T)) splits.

\paragraph{DAVIS~2017~\cite{pont20172017}.}
DAVIS~2017 is a widely adopted benchmark for video object segmentation featuring high-quality, pixel-level annotations across 30 semi-supervised validation sequences. We conduct evaluations on the full validation split.

\paragraph{SeCVOS~\cite{zhang2026advancingcomplexvideoobject}.}
SeCVOS is a benchmark for Semantic Complex Scenarios Video Object Segmentation, containing 160 manually annotated videos. It features challenging real-world conditions, including severe appearance changes, heavy occlusions, and dynamic scene variations. We evaluate on the full dataset.

\paragraph{MOSEv2~\cite{MOSEv2}.}
MOSEv2 is a large-scale benchmark designed for complex video object segmentation in crowded real-world environments. It contains 5{,}024 videos with 701{,}976 masks across 200 categories, emphasizing complex motion, disappearing objects, and long-term temporal dependencies. 

\paragraph{LVOS~\cite{hong2023lvos}.} 
LVOS is a benchmark for long-term video object segmentation, providing diverse and challenging scenarios for assessing long-range temporal tracking fidelity.

\paragraph{YouTube-VOS~\cite{youtube}.} 
We utilize the 2019 test split of YouTube-VOS, a large-scale benchmark for semi-supervised video object segmentation comprising 541 test videos across 65 seen categories and additional unseen categories to evaluate cross-category generalization. The evaluation set includes 1{,}092 unique object instances.

\subsection{Benchmark Setup}
\label{app:benchmark_setup}

\paragraph{Hardware Environments.}
Evaluations are conducted across three GPU hardware tiers: (1) NVIDIA A100-SXM4 (40\,GB) for ceiling throughput; (2) NVIDIA RTX 3090 (24\,GB) to evaluate OOM susceptibility under consumer limits; and (3) NVIDIA Tesla P100 (16\,GB) to test survival on memory-constrained legacy hardware.

\paragraph{Environment.}
All experiments are implemented in PyTorch with CUDA 12.6 using \texttt{torch.inference\_mode()}. Mixed precision (\texttt{torch.autocast}) is deployed with \texttt{bfloat16} (and \texttt{TF32}) on Ampere GPUs (A100, RTX 3090) and \texttt{float16} on Pascal (Tesla P100). Peak VRAM (\texttt{max\_memory\_reserved}) and energy consumption (NVML) are tracked across trials with GPU synchronization.

\subsection{Hardware Efficiency}
\label{apendix:hardware_scaling_combined}

\begin{figure}[t]
% \vspace{-26mm}
    \centering

    % (a) A100 Base+
    \begin{subfigure}{\linewidth}
        \centering
        \includegraphics[width=1.0\linewidth]{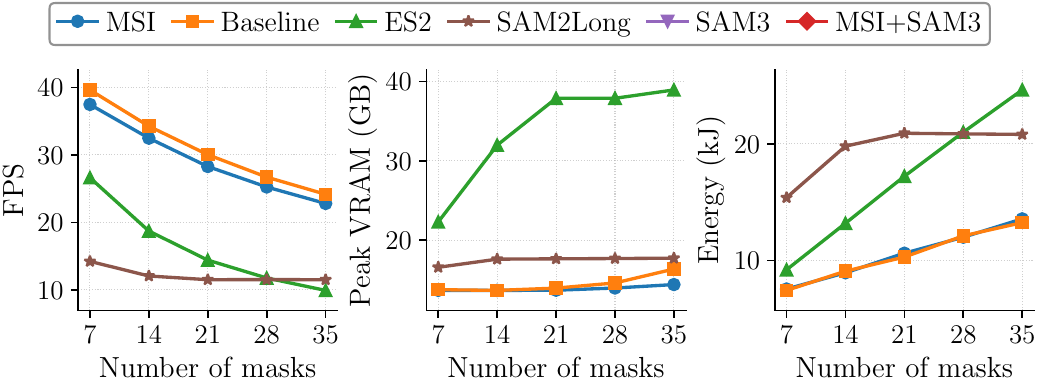}
        \vspace{-2mm}
        \caption{NVIDIA A100 (40\,GB) -- Base+ models}
        \label{fig:fps_a100_base}
    \end{subfigure}

    \vspace{1mm}

    % (b) A100 Large
    \begin{subfigure}{\linewidth}
        \centering
        \includegraphics[width=1.0\linewidth]{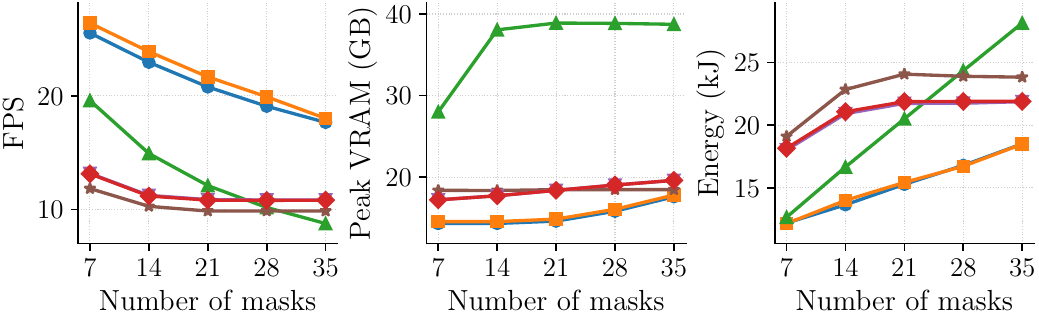}
        \vspace{-2mm}
        \caption{NVIDIA A100 (40\,GB) -- Large models}
        \label{fig:fps_a100_large}
    \end{subfigure}

    \vspace{1mm}

    % (c) RTX 3090 Large
    \begin{subfigure}{\linewidth}
        \centering
        \includegraphics[width=1.0\linewidth]{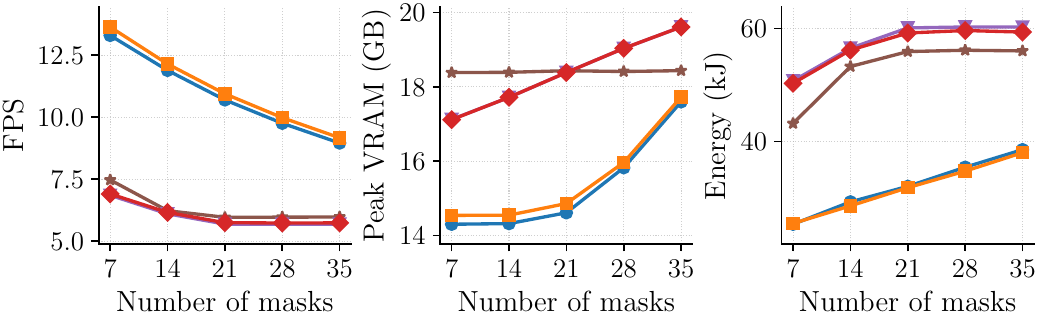}
        \vspace{-2mm}
        \caption{NVIDIA RTX 3090 (24\,GB) -- Large models}
        \label{fig:fps_3090_large}
    \end{subfigure}

    \vspace{-2mm}

    \caption{\textbf{Hardware scaling across backbones and devices.} Scaling profiles of throughput (FPS), peak VRAM (GB), and energy (kJ) across retrieval depths $M \in \{7, \dots, 35\}$: (a) A100 Base+; (b) A100 Large; (c) RTX 3090 Large.}
    \label{fig:hardware_scaling_combined}
    \vspace{-3mm}
\end{figure}

Figure~\ref{fig:hardware_scaling_combined} benchmarks throughput (FPS), peak reserved VRAM (GB), and energy (kJ) on NVIDIA A100 (40\,GB) and RTX~3090 (24\,GB) for Base+ and Large backbones ($M \in \{7, \dots, 35\}$).

\paragraph{NVIDIA A100.}
On the A100 GPU (Figures~\ref{fig:fps_a100_base} and \ref{fig:fps_a100_large}), MSI exhibits stable scaling, sustaining real-time throughput ($37.5 \to 22.8$\,FPS) and bounded energy ($7.56 \to 13.57$\,kJ) on Base+ models while matching baseline speed on Large backbones. $\text{MSI}+{\text{SAM3}}$ scales without latency penalty. In contrast, ES2 suffers severe resource inflation, with VRAM nearing physical capacity ($39.88$\,GB at $M=35$) and energy surging to $24.63$\,kJ.

\paragraph{NVIDIA RTX 3090.}
Figure~\ref{fig:fps_3090_large} evaluates Large models under the 24\,GB memory limit. While ES2 Large immediately crashes with Out-Of-Memory (OOM) at all levels ($M \ge 7$), MSI Large and $\text{MSI}+{\text{SAM3}}$ operate stably within physical VRAM limits without throughput degradation.

\subsection{Model-Agnostic}

To validate the generalizability of our proposed Methods (MSI) framework beyond modern Vision Transformer foundation models (such as SAM~2 and SAM~3), we integrate MSI into \textbf{XMem}~\cite{cheng2022xmem}: Long-Term Video Object Segmentation with an Atkinson-Shiffrin Memory Model.

\begin{table*}[t]
\centering
\caption{\textbf{Video Object Segmentation performance of MSI on XMem~\cite{cheng2022xmem}.} Comprehensive evaluation across six standard VOS benchmarks under varying target energy retention thresholds $\varepsilon \in \{0.999, 0.99, 0.95, 0.90\}$. The metric is $\mathcal{J}\&\mathcal{F}$ (\%). Throughput (FPS) is benchmarked on a single NVIDIA A100 GPU (40\,GB). \textbf{Bold} and \underline{underline} highlight the best and second-best scores, respectively.}
\label{tab:xmem_benchmark}
\setlength{\tabcolsep}{7.0pt}
\renewcommand{\arraystretch}{1.15}
\resizebox{\textwidth}{!}{%
\begin{tabular}{l ccccccc c}
\toprule
\textbf{Configuration} & \textbf{SA-V (T)} & \textbf{DAVIS '17} & \textbf{SA-V (V)} & \textbf{SeCVOS} & \textbf{LVOS} & \textbf{YTVOS} & \textbf{Avg}$\uparrow$ & \textbf{FPS$\uparrow$} \\
\midrule
XMem (Baseline)                  & \textbf{55.2} & \textbf{86.3} & \underline{58.6} & \underline{46.6} & \textbf{64.9} & \textbf{82.0} & \textbf{65.6} & \textbf{31.0} \\
\midrule
+ MSI ($\varepsilon = 0.999$)   & \underline{54.5} & \underline{85.7} & \textbf{59.2} & \textbf{46.7} & \underline{64.2} & \underline{81.8} & \underline{65.4} & 27.9 \\
+ MSI ($\varepsilon = 0.99$)    & 48.4 & 84.3 & 52.8 & 46.1 & 61.2 & 80.3 & 62.2 & \underline{30.7} \\
+ MSI ($\varepsilon = 0.95$)    & 16.5 & 46.3 & 20.4 & 35.2 & 25.7 & 47.5 & 31.9 & 29.0 \\
+ MSI ($\varepsilon = 0.90$)    & 11.4 & 27.9 & 10.0 & 27.3 & 14.0 & 24.7 & 19.2 & 28.5 \\
\bottomrule
\end{tabular}%
}
\vspace{-2mm}
\end{table*}

\paragraph{Experimental Setup.}
Following standard semi-supervised set up, we benchmark XMem on six standard datasets~\ref{app:dataset}

We apply MSI across all three memory components in XMem~\cite{cheng2022xmem} by compressing and subsequently restoring the representations to simulate low-rank memory reduction during inference:
\begin{itemize}[leftmargin=*]
    \item \textbf{Sensory Memory:} MSI is applied to the recurrent hidden state tensor and restored in-place.
    \item \textbf{Working Memory:} Incoming key and value tensors are compressed via MSI and restored prior to memory matching.
    \item \textbf{Long-Term Memory:} Consolidated prototype key and value representations are compressed using MSI and restored before long-term storage.
\end{itemize}

As reported in Table~\ref{tab:xmem_benchmark}, as $\varepsilon$ decreases, accuracy decreases monotonically across all benchmarks.
\begin{itemize}
    \item At $\varepsilon = 0.999$, MSI maintains near-baseline performance across all six benchmarks, achieving scores of $54.5$ on SA-V (T), $85.7$ on DAVIS '17, $59.2$ on SA-V (V), $46.7$ on SeCVOS, $64.2$ on LVOS, and $81.8$ on YTVOS. Notably, MSI slightly outperforms the baseline on SA-V (V) and SeCVOS by $0.6$ and $0.1$ points, respectively.
    \item At $\varepsilon = 0.99$, the average $\mathcal{J}\&\mathcal{F}$ score decreases to $62.2$ points, with the largest drops observed on SA-V (T) and SA-V (V), which fall to $48.4$ and $52.8$, respectively. In comparison, SeCVOS remains relatively robust at $46.1$, only $0.5$ points below the baseline, while DAVIS '17 and YTVOS retain scores of $84.3$ and $80.3$, respectively.
    \item At $\varepsilon = 0.95$, the average score drops substantially to $31.9$ points, a decrease of $33.7$ points relative to the baseline. SA-V (T) and SA-V (V) experience particularly severe degradation, reaching $16.5$ and $20.4$, respectively. Meanwhile, DAVIS '17 falls to $46.3$, LVOS to $25.7$, and SeCVOS to $35.2$. YTVOS exhibits the highest performance at this threshold, achieving $47.5$ points.
    \item At $\varepsilon = 0.90$, performance deteriorates further, with the average $\mathcal{J}\&\mathcal{F}$ score reaching only $19.2$ points. SA-V (V) records the lowest score at $10.0$, followed by SA-V (T) at $11.4$ and LVOS at $14.0$. DAVIS '17, SeCVOS, and YTVOS also decline to $27.9$, $27.3$, and $24.7$, respectively. 
\end{itemize}
These results demonstrate that overly aggressive energy truncation severely compromises the feature representations required for reliable video object tracking.

\subsection{Qualitative Results}
\label{ap:qualitative}

We evaluate qualitative tracking on YouTube-VOS across two challenging scenarios: fast dynamic motion and multi-target tracking.

\begin{figure*}[t]
    \centering
    \includegraphics[width=1.0\linewidth]{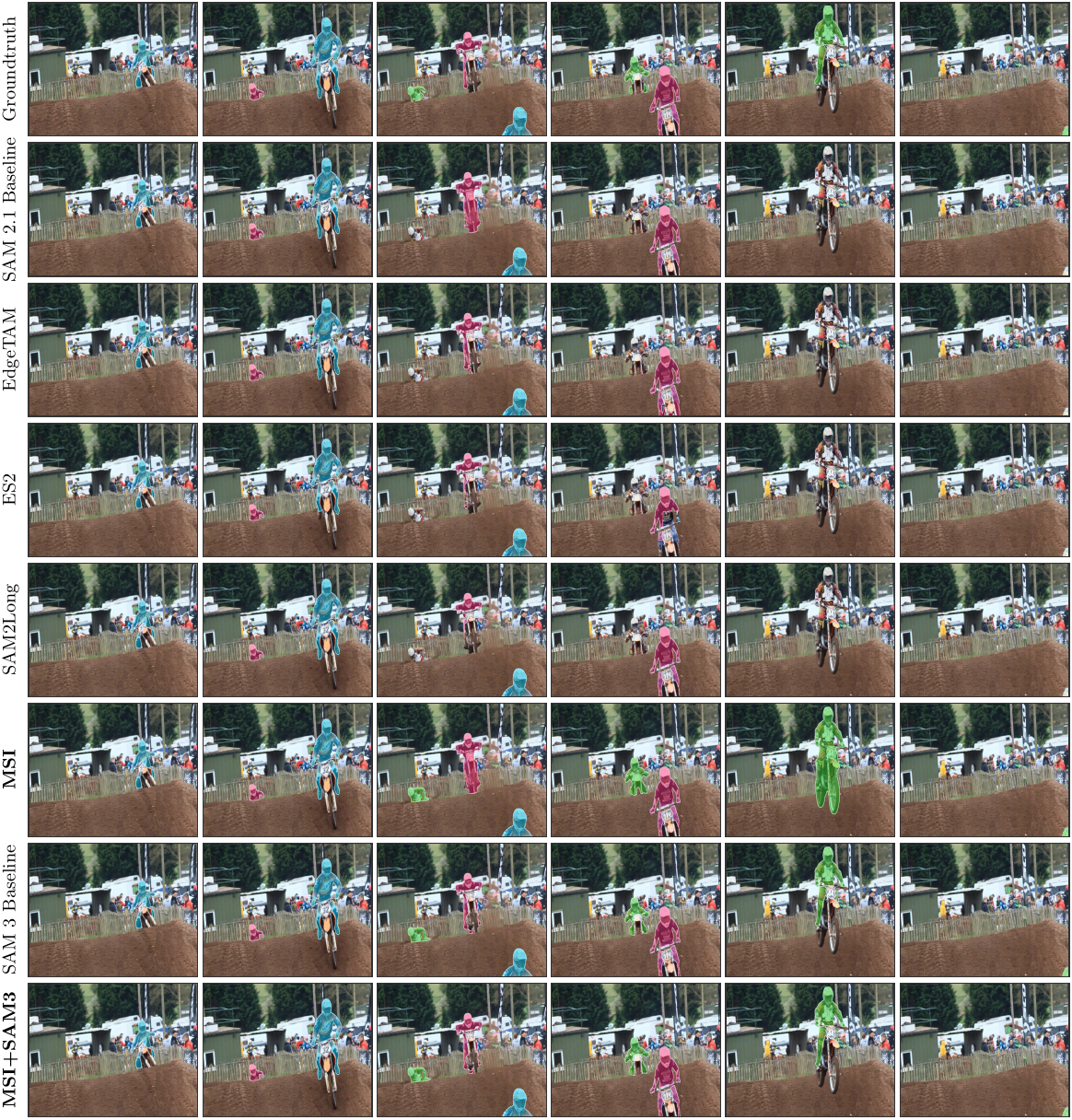}
    \vspace{-2mm}
    \caption{\textbf{Qualitative tracking under fast dynamic motion on YouTube-VOS.} Segmentation comparison across 6 consecutive keyframes. While baseline SAM~2, ES2, and EdgeTAM suffer from boundary drift, MSI maintains sharp and temporally consistent object boundaries.}
    \label{fig:ytvos_example1}
    \vspace{-3mm}
\end{figure*}

\paragraph{Fast Dynamic Motion.}
As shown in Figure~\ref{fig:ytvos_example1}, baseline SAM~2, ES2, and EdgeTAM suffer from boundary drift and mask degradation during rapid movement. In contrast, MSI maintains sharp object contours matching the Groundtruth without memory leakage.

\begin{figure*}[t]
    \centering
    \subfloat[Overall sequence tracking across 6 keyframes]{
        \includegraphics[width=0.98\linewidth]{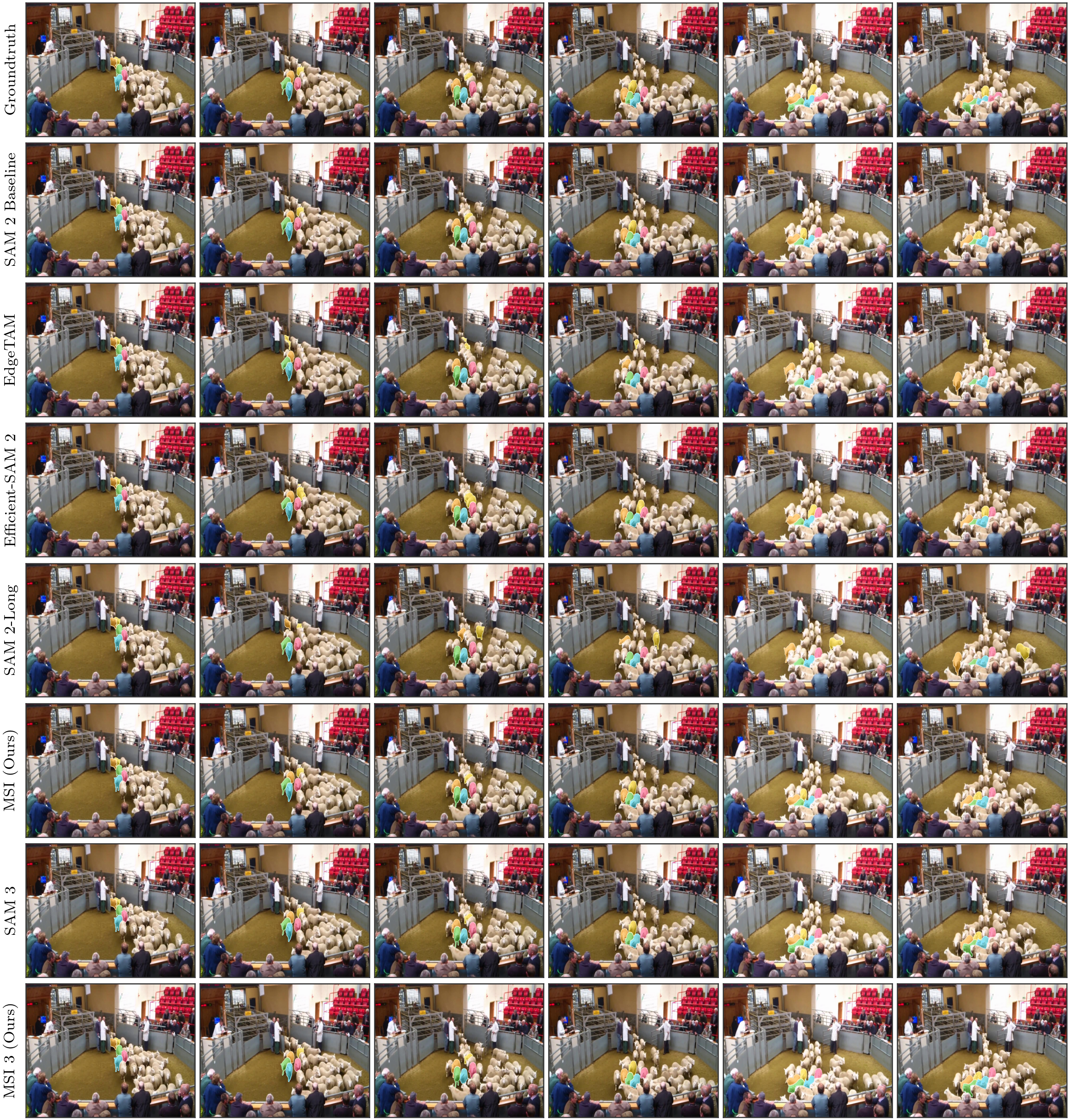}
        \label{fig:ytvos_ex3_seq}
    } 
    \vspace{0.1cm}
    \subfloat[Detailed zoom-in comparison at Frame 3]{
        \includegraphics[width=0.98\linewidth]{fig/vis/quality/figure_ytvos_example3_frame3_zoomin_1x8.pdf}
        \label{fig:ytvos_ex3_zoom}
    }
    \vspace{-2mm}
    \caption{\textbf{Multi-Target Tracking on YouTube-VOS.}}
    \label{fig:ytvos_example3_combined}
    \vspace{-3mm}
\end{figure*}

\paragraph{Multi-Target Tracking.}
Figure~\ref{fig:ytvos_example3_combined} illustrates tracking under crowded interactions. While competing methods (ES2 and SAM~2-Long) miss occluded targets in the Frame~3 close-up view (Figure~\ref{fig:ytvos_ex3_zoom}), MSI reliably separates each overlapping instance.

\begin{figure}[t]
% \vspace{12mm}
    \centering

    % ===================== HOSVD (a) =====================
    \begin{subfigure}{\linewidth}
        \centering

        \begin{subfigure}{0.24\linewidth}
            \centering
            \includegraphics[width=\linewidth]{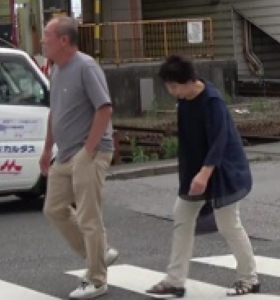}
            \subcaption*{$\varepsilon = 0.6$}
        \end{subfigure}\hfill
        \begin{subfigure}{0.24\linewidth}
            \centering
            \includegraphics[width=\linewidth]{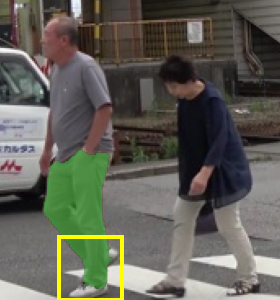}
            \subcaption*{$\varepsilon = 0.8$}
        \end{subfigure}\hfill
        \begin{subfigure}{0.24\linewidth}
            \centering
            \includegraphics[width=\linewidth]{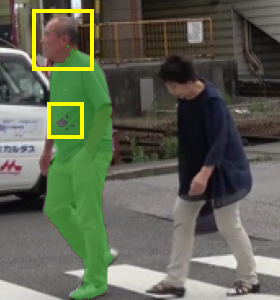}
            \subcaption*{$\varepsilon = 0.9$}
        \end{subfigure}\hfill
        \begin{subfigure}{0.24\linewidth}
            \centering
            \includegraphics[width=\linewidth]{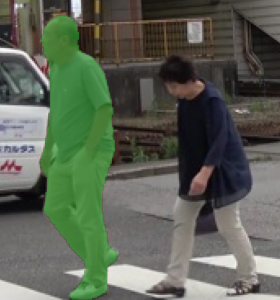}
            \subcaption*{$\varepsilon = 0.99$}
        \end{subfigure}

        % \vspace{-1mm}
        \subcaption*{(a) HOSVD}
    \end{subfigure}

    \vspace{1mm}

    % ===================== MSI (b) =====================
    \begin{subfigure}{\linewidth}
        \centering

        \begin{subfigure}{0.24\linewidth}
            \centering
            \includegraphics[width=\linewidth]{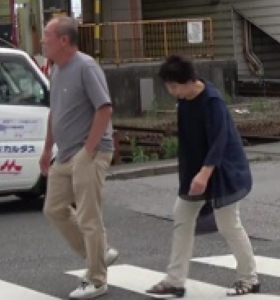}
            \subcaption*{$\varepsilon = 0.6$}
        \end{subfigure}\hfill
        \begin{subfigure}{0.24\linewidth}
            \centering
            \includegraphics[width=\linewidth]{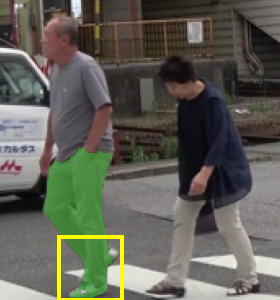}
            \subcaption*{$\varepsilon = 0.8$}
        \end{subfigure}\hfill
        \begin{subfigure}{0.24\linewidth}
            \centering
            \includegraphics[width=\linewidth]{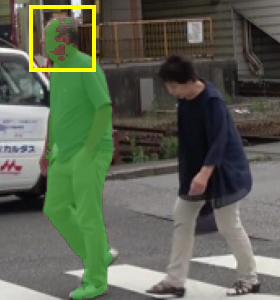}
            \subcaption*{$\varepsilon = 0.9$}
        \end{subfigure}\hfill
        \begin{subfigure}{0.24\linewidth}
            \centering
            \includegraphics[width=\linewidth]{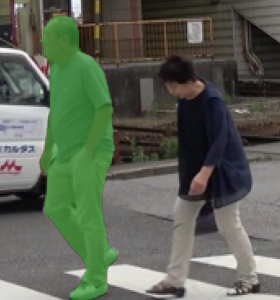}
            \subcaption*{$\varepsilon = 0.99$}
        \end{subfigure}

        % \vspace{-1mm}
        \subcaption*{(b) MSI}
    \end{subfigure}

    \vspace{-2mm}
    \caption{\textbf{Qualitative mask reconstruction comparison across variance thresholds $\varepsilon$.}}
    \label{fig:hosvdvsmsivisaual}
    \vspace{-3mm}
\end{figure}

\paragraph{MSI vs. HOSVD.} Fig.~\ref{fig:hosvdvsmsivisaual} further visualizes the predicted masks produced by HOSVD and MSI under different $\varepsilon$ values on SAM~2.1-B+. As $\varepsilon$ increases from $0.6$ to $0.99$, both methods progressively recover more complete object masks. Notably, MSI better recovers a missing region at $\varepsilon=0.9$, while both methods recover complete masks at $\varepsilon=0.99$. 

\newpage

\end{document}